%% file: template.tex
\documentclass{article}

\usepackage{arxiv}

\usepackage{doi}

\usepackage{times}  
\usepackage{helvet}  
\usepackage{courier}  
\usepackage{graphicx} 
\usepackage{url}  
\usepackage{algorithm}
\usepackage{algorithmic}
\usepackage{subcaption}
\usepackage{hyperref}
\usepackage{amsmath}
\usepackage{cleveref}
\usepackage{amsfonts,mathrsfs}
\usepackage{booktabs} 

\usepackage{xr}
\usepackage{newfloat}

\DeclareMathOperator*{\argmax}{arg\,max}
\title{Towards Reasoning for PDE Foundation Models: A Reward-Model-Driven Inference-Time-Scaling Algorithm}

\author{ \href{https://orcid.org/0009-0000-9015-7041}{\includegraphics[scale=0.06]{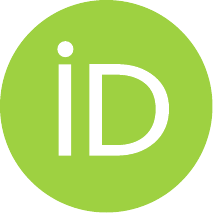}\hspace{1mm}Siddharth Mansingh}\thanks{Corresponding Author: smansingh@lanl.gov}\\
	\And
	\href{https://orcid.org/0000-0001-5080-7038}{\includegraphics[scale=0.06]{orcid.pdf}\hspace{1mm}James Amarel}  \\
    \And
	\href{https://orcid.org/0009-0004-9808-1693}{\includegraphics[scale=0.06]{orcid.pdf}\hspace{1mm}Ragib Arnab} \\
    \And
	\href{https://orcid.org/0000-0002-9434-7691}{\includegraphics[scale=0.06]{orcid.pdf}\hspace{1mm}Arvind Mohan}\thanks{Corresponding Author: arvindm@lanl.gov} \\
    \And
	\href{https://orcid.org/0000-0002-2833-7903}{\includegraphics[scale=0.06]{orcid.pdf}\hspace{1mm}Kamaljeet Singh} \\
    \And
	\href{https://orcid.org/0000-0002-4672-9484}{\includegraphics[scale=0.06]{orcid.pdf}\hspace{1mm}Gerd J. Kunde} \\
    \And
	\href{https://orcid.org/0000-0002-4157-134X}{\includegraphics[scale=0.06]{orcid.pdf}\hspace{1mm}Nicolas Hengartner} \\
    \And
	\href{https://orcid.org/0000-0001-6812-9933}{\includegraphics[scale=0.06]{orcid.pdf}\hspace{1mm}Benjamin Migliori} \\
    \And
	\href{https://orcid.org/0000-0003-2909-5704}{\includegraphics[scale=0.06]{orcid.pdf}\hspace{1mm}Emily Casleton} \\
    \And
	\href{https://orcid.org/0000-0002-5593-9205}{\includegraphics[scale=0.06]{orcid.pdf}\hspace{1mm}Nathan A. Debardeleben} \\
    \And
	\href{https://orcid.org/0000-0002-7500-145X}{\includegraphics[scale=0.06]{orcid.pdf}\hspace{1mm}Ayan Biswas} \\
    \And
	\href{https://orcid.org/0000-0002-1353-3688}{\includegraphics[scale=0.06]{orcid.pdf}\hspace{1mm}Diane Oyen} \\
    \And
	\href{https://orcid.org/0000-0002-6473-1887}{\includegraphics[scale=0.06]{orcid.pdf}\hspace{1mm}Earl Lawrence} \\
    \\
    Los Alamos National Laboratory, Los Alamos, NM, US
}

\hypersetup{
pdftitle={Towards Reasoning for PDE Foundation Models: A Reward-Model-Driven Inference-Time-Scaling Algorithm},
pdfsubject={physics.comp-ph,cs.LG},
pdfauthor={Siddharth Mansingh, James Amarel, Ragib Arnab, Arvind Mohan, Kamaljeet Singh, Gerd J. Kunde, Nicolas Hengartner, Benjamin Migliori, Emily Casleton, Nathan A. Debarledeben, Ayan Biswas, Diane Oyen, Earl Lawrence},
pdfkeywords={PDE, Reasoning, Surrogate Modeling, Foundational Models, Test Time Compute},
}

\begin{document}
\maketitle

\begin{abstract}
	Partial Differential Equations (PDEs) are the bedrock for modern computational sciences and engineering, and inherently computationally expensive. While PDE foundation models have shown much promise for simulating such complex spatio-temporal phenomena, existing models remain constrained by the pretraining datasets and struggle with auto-regressive rollout performance, especially in out-of-distribution (OOD) cases. Furthermore, they have significant compute and training data requirements which hamper their use in many critical applications. Inspired by recent advances in ``thinking" strategies used in large language models (LLMs), we introduce the first test-time computing (TTC) strategy for PDEs that utilizes computational resources during inference to achieve more accurate predictions with fewer training samples and smaller models. We accomplish this with two types of reward models that evaluate predictions
of a stochastic based model for spatio-temporal consistency.
We demonstrate this method on compressible Euler-equation simulations from the PDEGym benchmark and show that TTC captures improved predictions relative to standard non-adaptive auto-regressive inference.
This TTC framework marks a foundational step towards more advanced reasoning algorithms or PDE modeling, inluding building reinforcement-learning-based approaches, potentially transforming computational workflows in physics and engineering.
\end{abstract}


\section*{Introduction}
\label{intro}
Scientific Foundation Models (FMs) have rapidly emerged as powerful computational engines across diverse scientific and engineering disciplines, particularly by accelerating simulations involving Partial Differential Equations (PDEs) \cite{hao2024dpot,ye2024pdeformer,herde2024poseidon}. Learning PDE solution operators directly from data and physics-based constraints, these models promise substantial computational savings over traditional numerical methods, such as finite-volume or spectral approaches \cite{long2018pde,raissi2018hidden}. This efficiency opens unprecedented possibilities for rapid scientific forecasting, real-time predictions, and iterative engineering workflows. Yet, despite these significant advances, two fundamental challenges persist: a) PDE-based FMs frequently suffer from compounding errors and distribution shifts when performing long time-horizon, autoregressive rollouts. This vulnerability is particularly pronounced in out-of-distribution (OOD) scenarios, which frequently arise in practical, safety-critical contexts and severely impact reliability \cite{gupta2023towards,lippe2023pderefiner}. b) Another critical bottleneck is the current heavy reliance on extensive fine-tuning datasets for downstream applications. In practice, most real-world applications suffer from a lack of high quality downstream data, as acquiring them may be impractical or prohibitively expensive.

Addressing these persistent issues calls for fundamentally rethinking how computational resources are efficiently leveraged in Scientific FMs. This is particularly significant in computational sciences where we have access to high fidelity simulations that can address downstream problems, but are prohibitively expensive to run. Therefore, unlike many LLM use cases, the scientific FM solution \textit{must be significantly more efficient}, and this places a practical upper bound on how much compute power ML methods can expend, both in training and inference. One promising direction to achieve this is inspired by recent advances in reasoning-based strategies from the field of LLMs—such as Chain-of-Thought (CoT) \cite{wei2022chain} and Tree-of-Thought (ToT) \cite{yao2023tree}. These methods deliberately allocate more computational resources during inference time to internally evaluate, self-correct, and dynamically adapt model predictions without additional training data, significantly boosting performance even under challenging distribution shifts. A key outcome has been the competitive performance of reasoning LLMs with an order of magnitude fewer parameters than their non-reasoning counterparts, and often with lesser training data. Therefore, this possibility of getting higher accuracy with smaller models and lesser training data is extremely attractive to the computational sciences community. 

Consequently, there have been increasing conversations if LLM reasoning-inspired inference methods can be meaningfully adapted to the rigorous and physically constrained context of scientific modeling~\cite{duraisamy2025active}. A special focus is on computational problems where governing PDEs are available and are mission critical to many applications. By analogy to LLM reasoning, reasoning in PDE FMs can be interpreted as systematically employing inference-time computation to evaluate, compare, and select among multiple plausible candidate solutions, guided by a reward signal. However, this analogy should be carefully scrutinized. How closely does computational “reasoning” mirror human or classical algorithmic reasoning strategies in scientific disciplines? Additionally, while LLM strategies provide a useful starting analogy, significant differences in the nature of problems and constraints in PDEs necessitate careful re-evaluation of how these reasoning strategies might be appropriately adopted.

To initiate exploration of these questions, we introduce a novel test-time computation (TTC) approach explicitly tailored for PDE foundation models: to our knowledge the first of its kind in this domain.  We attempt to articulate precisely what “reasoning” should mean in the context of PDEs for scientific and engineering problems. Our approach explores the possibility of dynamically generating multiple candidate solutions at each inference step and evaluating their relative quality using a reward model, ultimately selecting the most promising candidate. We present a framework that uses beam-search style sampling, so as to explore the performance gains that can be made with solely reward-model driven self evaluation~\cite{xie2023self,zhu2024deductive} without more expensive components like reinforcement learning (RL). As such, these results are a foundational first step towards fully adaptive, RL-based reasoning algorithms, much like the early era of LLM reasoning models~\cite{snell2024scaling}. To this end, we experiment with two distinct reward-modeling strategies: Analytical Reward Models (ARMs), grounded in explicit physical constraints for interpretability, and learned Process Reward Models (PRMs), which provide greater flexibility at the potential cost of reduced interpretability. Importantly, the proposed method specifically addresses the realistic scenario where downstream fine-tuning data is sparse or limited. 

We emphasize that this paper does not claim a definitive implementation of reasoning for PDE foundation models. Instead, it shows the promise of test time compute. By evaluating performance on nontrivial PDE benchmarks, specifically nonlinear compressible Euler equations in the PDEGym environment \cite{herde2024poseidon}, and considering realistic scenarios with severely limited downstream data, we provide concrete initial conditions, baselines, and insights for future research.

Here we summarize the novel contributions of our work:
\begin{itemize}
    \item We introduce a novel TTC-based reasoning framework for PDE FMs that achieves state-of-the-art downstream accuracy after fine-tuning on only 6.25\% of the training data required by an equivalent baseline FM without TTC. This demonstrates significant gains in sample efficiency. This is critical for PDE applications, where labeled data generation can be prohibitively expensive.
    \item Our approach achieves these results using a foundation model comprised of only about 5 million parameters: a significant reduction compared to leading PDE models on identical benchmarks, which range from 21M to 0.7 billion parameters~\cite{herde2024poseidon,hao2024dpot}. Thus, our methodology delivers not only exceptional sample efficiency but also parameter and computational efficiency, highlighting the feasibility of building compact yet powerful foundation models.
    \item A  key innovation in our TTC paradigm is the introduction of learned Process Reward Models (PRMs) alongside analytical reward functions. Notably, these PRMs, matched in size to the FM itself, are effectively trained using only 12.5\% of the original training samples, further underscoring our method’s computational efficiency. 
\end{itemize}


\begin{figure*}[h]
    \centering
    \includegraphics[width=1\textwidth]{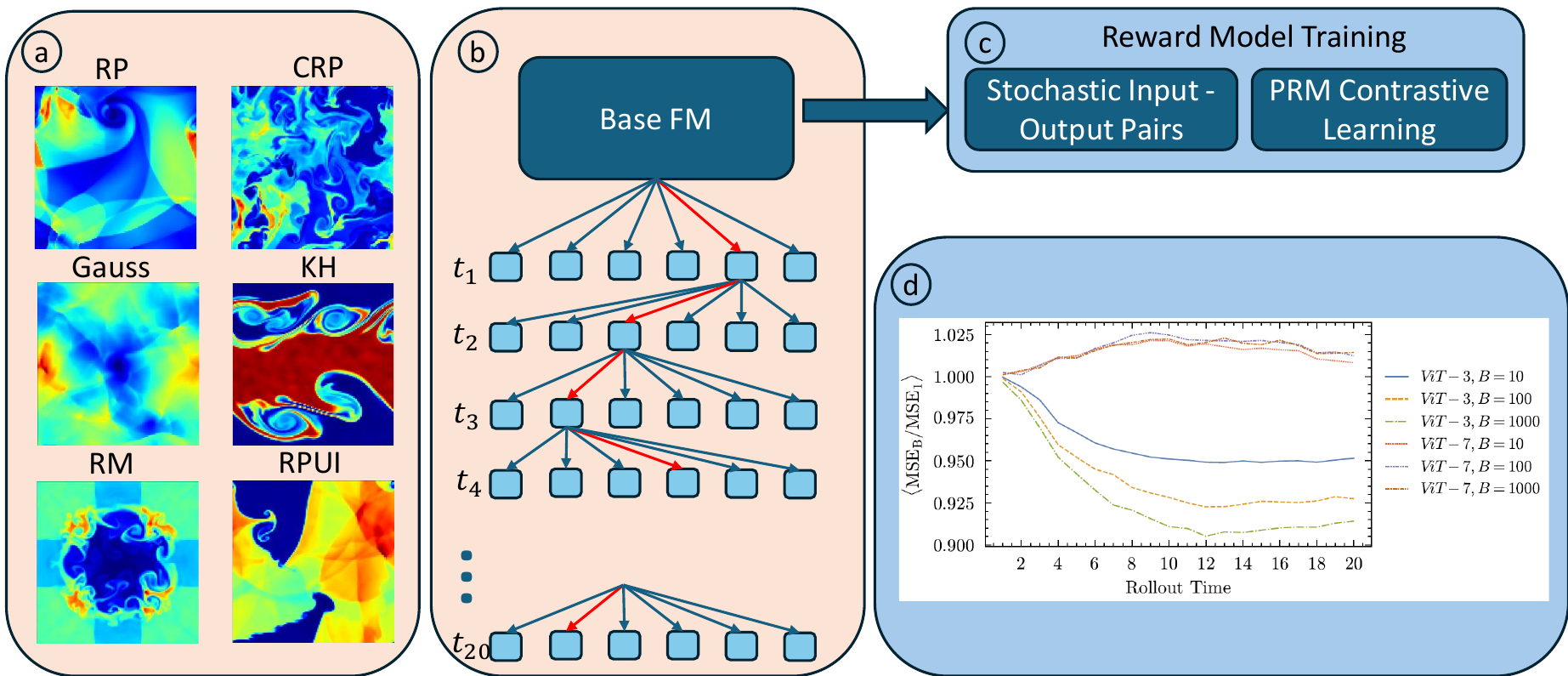}
    \caption{a) Snapshots of Compressible Euler PDE dataset used for pretraining (RP, CRP, Gauss and KH) and downstream tasks (RM and RPUI). b) Greedy selection strategy to select the prediction with the best reward from a set of candidate predictions generated by the base foundational model. c) Outputs of the pretrained model are used for training the process reward model (PRM) using contrastive learning. d) Test Time Compute performance on CRP pretraining task. Reward Model Driven TTC provides substantial gains in MSE.}
    \label{fig:ttc_summary}
\end{figure*}

\begin{table*}[ht!]
\centering
\caption{Percentage Sample Gain $\left(1-\langle\text{MSE}_{\text{B}=1000}/\text{MSE}_{\text{B}=1}\rangle\right)$ comparison of various models under different reward models. Process-Reward Models perform better across the board for both pretraining and finetuning datasets.}
\label{table:model_comparison}
\begin{tabular}{lccc ccc }
\toprule
& \multicolumn{3}{c}{Analytical Reward Model} & \multicolumn{3}{c}{Process Reward Model}  \\
\cmidrule(lr){2-4} \cmidrule(lr){5-7}
& ViT-3 & ViT-5 & ViT-7 & ViT-3 & ViT-5 & ViT-7  \\
\midrule\midrule
CE-CRP   &   8.582  & 11.980  &  -1.427          &  14.289      &  14.147      & 13.225           \\
\midrule
CE-RP  & 0.221  &  1.342      & 1.857        &    17.716    &  22.488      &        22.491  \\
\midrule
CE-Gauss & -0.220       &    -2.365    &    2.276    &    20.591    & 22.071       &  22.714          \\
\midrule
CE-KH    &  -1.206      &   -0.711     &    -7.385    &   5.088     &  3.683      &   -0.455      \\
\midrule
CE-RM    &   0.100     &   -0.521     &   -0.317     &  3.421      & 3.765 &   3.639             \\
\midrule
CE-RPUI  &   16.919     & 11.99   & 7.327        &  16.215       & 18.197       &  25.722         \\
\bottomrule
\end{tabular}
\end{table*}
\section*{Related Work and Proposed Approach}
\label{sec:litreview}
The current frontier of large language model (LLM) reasoning research faces a fundamental bottleneck: the scarcity of high-quality training data, particularly in the form of detailed, step-by-step reasoning trajectories. Human annotation of such trajectories is prohibitively expensive and time-consuming, limiting the scalability of supervised reasoning approaches \cite{lightman2023letsverifystepstep}. To overcome this challenge, recent advances have shifted toward LLM-driven search algorithms that leverage external verification to automatically generate reasoning traces through trial-and-error exploration \cite{luo2024trial}. These generated traces are then used to train Process Reward Models (PRMs), which enable models to learn the reasoning process itself, rather than just the final answers \cite{zhang2024restmcts}. This forms a ``reinforced cycle" that integrates search and learning, aligning with Richard Sutton’s ``bitter lesson" prediction that scalable AI systems will rely heavily on computation rather than handcrafted knowledge \cite{sutton2019bitter}. Importantly, this approach scales reasoning capabilities by increasing train-time compute and search, rather than merely expanding model parameters. Complementary work suggests that scaling test-time compute via search and planning may unlock more reasoning ability than parameter scaling alone \cite{snell2025scaling}. Furthermore, reinforcement learning (RL) is emerging not only as a mechanism for fine-tuning models to align with human values following the Reinforcement Learning from Human Feedback (RLHF) paradigm \cite{ouyang2022training} but also as a tool for data construction, trajectory selection, and model training in reasoning domains. This combined RL-driven data generation, train-time optimization, and test-time search represents a promising path toward LLMs capable of advanced, scalable reasoning.

\subsection*{An approach for reasoning in PDE FMs}


While reasoning in contemporary LLMs offers an iterative way of improving performance through local changes that provide some interpretability, similar local reward signals are harder to construct in PDE simulations. In prior works on PDE refinement, a notion of local reward is still missing \cite{lippe2023pderefiner}. At the heart of the TTC approach presented in this paper is the extraction of global signal from physical conservation laws or from a learnable surrogate loss function. Using a global reward signal, one should be able to discriminate samples that ``better'' adhere to physical conservation laws such as mass/momentum conservation, given a batch of plausible predictions. Choosing the ``best'' sample at each timestep of a autoregressive rollout with a greedy strategy helps the full trajectory predictions adhere globally to the conservation laws. However, this depends on the quality of the reward function, as would be made clear in the results.

\textbf{Problem Statement}: Given a generic time-dependent PDE in the function space $\mathscr{X} \in L\subset [D;\mathbb{R}^n]$ with initial condition $a\in \mathscr{X}$ and $\mathcal{L}$ and $B$ being the underlying differential and boundary operators, the task is to approximate the solution $u(t,x)$ with an approximate solution operator $\mathcal{S}: [0,T]\times \mathscr{X}\to \mathscr{X}$ where $u(t,x)$ obeys

\begin{equation} \label{eq:generic_pde}
    \partial_t u(x,t) + \mathcal{L}(u,\nabla_x u,\nabla_x^2 u,\cdots ) = 0, 
\end{equation}
with boundary conditions
\begin{align} \label{eq:generic_pde_boundary}
    \mathcal{B}(u)=0, \quad \forall (x,t)\in \partial D \times (0,T), \quad u(0,x) = a(x). \notag
\end{align}
In other words, the formulated solution operator can predict the state of the system at next timestep $u(t+1,x)$ given the current timestep $u(t,x)$. 
We learn the solution operator $S$ in a data driven manner by minimizing the difference between the predictions of the neural operator $S_\theta$ and ground truth training data which is present in the form of trajectories ${S(t,u_t,a_i)}$ for discrete times $t\in [0,1,\cdots, T]$ and initial conditions $a_i,\, i\in[1,N]$. Here $u_t$ is shorthand for the solution $u(t,x)$ of \ref{eq:generic_pde} corresponding to the initial condition $u(0,x)=a_i(x)$.The loss function we use for training 
\begin{equation}
    \mathcal{L}_\theta = \frac{1}{N(T+1)}\sum_{i=1}^N \sum_{t=0}^T\| S_\theta (t,u_t,a_i) - S(t,u_t,a_i)\|^p_{L^p(D)}
\end{equation}
only focuses on the consecutive timestep pairs in any given trajectory. Once the neural operator , given the state $u(t,x)$ corresponding to the initial condition $a_i$ and time $t$, the learned neural solution operator returns $u(t+1,x)$.
\begin{figure*}[ht]
    \begin{subfigure}[t]{0.5\textwidth}
        \includegraphics[width=\textwidth]{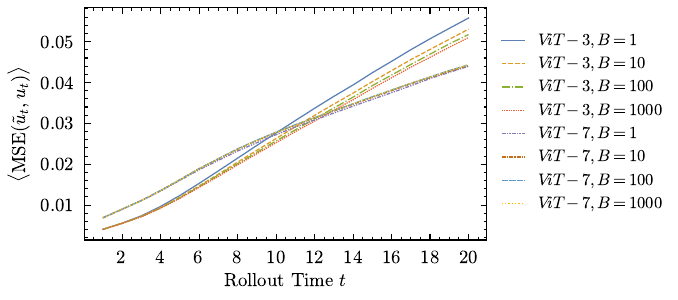}
        \caption{}
        \label{fig:crp_mass_maintext}
    \end{subfigure}
    \hfill
    \begin{subfigure}[t]{0.5\textwidth}
        \includegraphics[width=\textwidth]{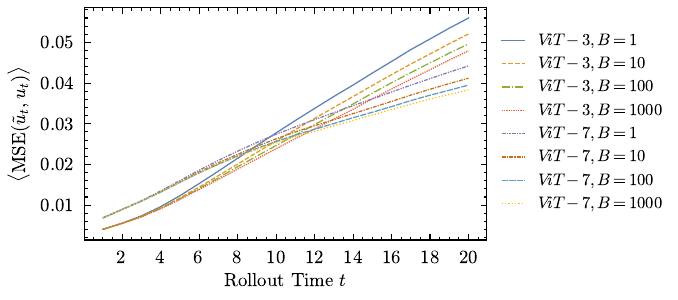}
        \caption{}
        \label{fig:crp_prm_maintext}
    \end{subfigure}
    
    \caption{Rollout performance of Greedy Selection Strategy on the CRP dataset, for \protect{\subref{fig:crp_mass_maintext}} Analytical Reward Model and \protect{\subref{fig:crp_prm_maintext}} Process-Reward Model. As the branching factor $B$ is increased, the error between predictions and ground truth reduce for all rollout times. PRM provides signficant improvements on error reduction, compared to ARM.}
    \label{fig:composite}
\end{figure*}


\section*{Dataset and Methodology}
\label{sec:methods}
The PDEGym dataset~\cite{herde2024poseidon} provides a standardized and comprehensive suite of benchmark data designed specifically for developing and evaluating operator-learning models across various PDEs. PDEGym notably distinguishes itself by aggregating diverse PDEs, such as the Navier-Stokes and compressible Euler (CE) equations, simulated using high-resolution numerical solvers including high-order spectral and finite-volume methods. In this study, we specifically focus on the CE subset from PDEGym, which presents complex nonlinear inviscid fluid dynamics characterized by multi-scale phenomena including shocks, rarefactions, and vortex structures. This subset includes multiple challenging scenarios, such as Riemann problems, Kelvin-Helmholtz instabilities, and Gaussian perturbations, each emphasizing distinct nonlinear dynamics and solution behaviors. All CE scenarios in PDEGym are defined on a standardized spatial domain, typically the two-dimensional unit square \(D = [0, 1]^2\), with periodic boundary conditions.
Each dataset provides comprehensive solution trajectories that comprise snapshots of fluid state variables—density (\(\rho\)), horizontal velocity (\(v_x\)), vertical velocity (\(v_y\)), and pressure (\(p\)) recorded at multiple uniformly spaced time instances. Specifically, trajectories are simulated from an initial state up to a fixed terminal time \(T = 1\), with snapshots stored at 21 discrete time points uniformly spaced within the interval, yielding rich temporal data essential for effective operator learning. 
In this study, we trained a based model produce accurate and stable auto-regressive rollouts from unique, unseen initial conditions (ICs) to the full 21 timestep trajectory and then improved upon those rollouts with our TTA approach, which adaptively allocates computational resources during inference. In this study, we specifically utilize four datasets from PDEGym for pretraining: CE-RP, CE-CRP, CE-KH, and CE-Gauss. For downstream, we use the CE-RPUI and CE-RM dataset, which have qualitatively different physics (Fig.~\ref{fig:ttc_summary}a), making them challenging downstream tasks.
\subsection*{Foundation Model Design}
Our architecture is based on the vision transformer (ViT) image-translation model, that has been well established in literature and consists of three main components: a \textbf{patch embedding encoder}, a \textbf{transformer bottleneck}, and a \textbf{patch reconstruction decoder}.
The encoder projects the input image into a sequence of non-overlapping patches using a convolutional layer with kernel size and stride equal to the patch size. Each patch is linearly mapped to a high-dimensional embedding space. Positional embeddings are added to retain spatial information.
Let $x \in \mathbb{R}^{B \times C \times H \times W}$ denote the input image. The encoder produces
\begin{equation}
z_0 = \text{Proj}(x) + E_{\text{pos}} \in \mathbb{R}^{B \times N \times D},
\end{equation}
where $N = \frac{H \cdot W}{P^2}$ is the number of patches of size $P^2$ and $D$ is the embedding dimension. The underlying periodic boundary conditions imposed in the dataset are reflected in the model by imposing circular padding. Furthermore since odd shaped kernels are better mathematical descriptors of PDE operators~\cite{long2018pde,long2019pde,mohan2023embedding}, we split each input image into odd-shaped patches i.e., $3X3,\, 5X5$. Thus we have three different kinds of ViTs corresponding to the patch size used, referred to as ViT-3, ViT-5 and ViT-7.

The encoded patch sequence is passed through a stack of standard Transformer encoder blocks, each consisting of multi-head self-attention (MHSA), layer normalization (LN), and a feed-forward network (FFN). This stage captures long-range interactions between distant patches, facilitating global context modeling.

Each Transformer block computes
\begin{equation}
z_{l+1} = z_l + \text{MHSA}(\text{LN}(z_l)) + \text{FFN}(\text{LN}(z_l)), \quad l = 1, \ldots, L,
\end{equation}
where $L$ is the number of Transformer layers.
The final sequence of embeddings is linearly projected back into pixel space using a learnable projection, followed by spatial rearrangement to form the output image. Specifically, each token is reshaped into a spatial patch and assembled to reconstruct the full image
\begin{equation}
\hat{x} = \text{Reshape}(\text{Proj}^{-1}(z_L)) \in \mathbb{R}^{B \times C \times H \times W}
\end{equation}
This decoder enables end-to-end image translation by reconstituting the spatial layout from the processed patch sequence.  Details on hyperparameters, computational resources and performance of the base models on the test set of pretraining datasets  have been provided in Supplementary Section.  

\textbf{Stochasticity}: A key difference from reasoning for LLMs that needs rethinking in PDE FMs, is the use of model stochasticity to generate multiple predictions for the same input. In LLMs, the model is probabilistic by default, due to internal tokenization and one-hot-encoding. This is not the case in most ML models for PDEs, as they are often deterministic in nature. Therefore, to utilize techniques like beam-search, we must explicitly introduce inherent stochasticity in the base PDE FM, without compromising the overall accuracy of the model.  A straightforward way to introduce stochasticity to any model is through the dropout mechanism. Instead of disabling dropout after training, as is the standard in machine learning, we keep dropout active during inference time, such that the FM produces a different prediction for the same input by sampling different dropout masks.

\subsection*{Downstream finetuning}

To finetune our base model, we start with the pretrained model and given any downstream task, we randomly select $n$ trajectories in the training set. The entire set of trainable parameters is then initialized by transferring the parameters from the pretrained model and then trained. Results for the base performance of finetuned models on different downstream tasks and various trajectories $n$ that were used for finetuning, without any reasoning, are shown in Supplementary Material.
\begin{figure*}[ht]
    \begin{subfigure}[t]{0.5\textwidth}
        \includegraphics[width=\textwidth]{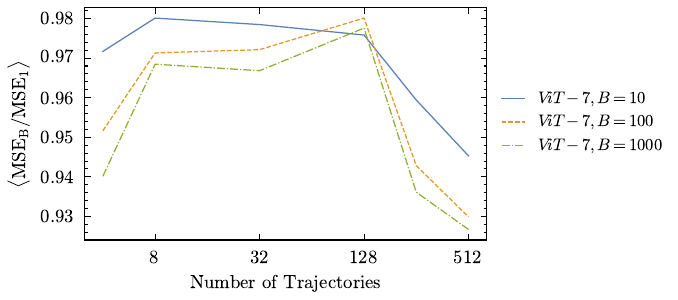}
        \caption{}
        \label{fig:rpui_mass_ratio_maintext}
    \end{subfigure}
    \hfill
    \begin{subfigure}[t]{0.5\textwidth}
        \includegraphics[width=\textwidth]{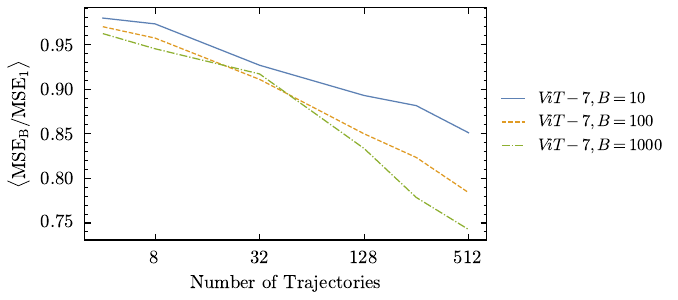}
        \caption{}
        \label{fig:rpui_prm_ratio_maintext}
    \end{subfigure}
    
    \caption{Sample Gain Ratio for ViT-7 model on downstream RPUI task, when finetuned over different number of trajectories using \protect\subref{fig:rpui_mass_ratio_maintext} ARM and \protect\subref{fig:rpui_prm_ratio_maintext} PRMs produce monotonic improvement of MSE on a per-sample basis as the models are trained on increasing number of trajectories.}
    \label{fig:finetune_ratio}
\end{figure*}
\subsection*{Reward Model}
\textbf{Analytic Reward Model}: Since the PDEs present in our datasets (both pretraining and finetuning) conserve mass, momentum and energy, as there is no external source for the mentioned quantities, it is possible to devise reward functions that favor solutions with mass/momentum/energy conservation. The mass conservation reward function can be written as 
\begin{equation} \label{eq:mass_cons}
\begin{aligned}
    &\text{ARM}_{\text{Mass}}(\tilde{u}(t),\tilde{u}(t+1))\\
    &\quad=  \frac{-|\sum_x \tilde{\rho}(t+1,x)-\sum_x\tilde{\rho}(t,x)|}{\sum_x\tilde{\rho}(t,x)}
    \end{aligned}
\end{equation}
While Eq \ref{eq:mass_cons} serves as the reward model that chooses samples with better mass conservation i.e., higher reward values imply better mass conservation, where a reward value of 0 corresponds to ideal mass conservation, the output of the reward model can also serve as a metric to decide how well a trajectory has conserved mass. 

Similarly, we define momentum conservation reward function 
\begin{align} \label{eq:momentum_cons}
    &\text{ARM}_{\text{Momentum}}(\tilde{u}(t),\tilde{u}(t+1))\\ = 
    &\frac{-\left|\sum_x \tilde{v}_x(t+1)\tilde{\rho}(t+1,x)-\sum_x\tilde{v}_x(t)\tilde{\rho}(t,x)\right|}{\left|\sum_x\tilde{v}_x(t)\tilde{\rho}(t,x)\right|} \notag
\end{align} 
and the energy conservation reward function 
\begin{align} \label{eq:energy_cons}
    &\text{ARM}_{\text{Energy}}(\tilde{u}(t),\tilde{u}(t+1))\\ =  &\frac{-|\sum_x \tilde{E}(t+1,x)-\sum_x\tilde{E}(t,x)|}{\sum_x\tilde{E}(t,x)}\notag.
\end{align}



\noindent\textbf{Process-Reward Model} In addition to the analytical reward functions, we train a learnable process-reward model (PRM) \cite{lightman2023letsverifystepstep} on the outputs of our foundation model. The PRM provides a scalar-valued score grading the quality of the next snapshot prediction given the current snapshot. To train the PRM, we sample 100 next-step predictions per initial condition for a pretrained/finetuned model (more details are included in the results section). The samples are ranked based on a chosen metric, in our case, MSE, against the ground truth. We select and save the triplet samples that correspond to predictions with maximum, median and minimum scores. Then, we incorporate a contrastive triplet margin loss to train the PRM: 

\begin{align}
\mathcal{L_{\text{PRM}}} = &\max(0, r_{\text{min}} - r_{\text{median}} + \alpha) \\  &\quad + \max(0, r_{\text{median}} - r_{\text{max}} + \alpha)\notag,
\end{align}
where $r_{\text{min}}$, $r_{\text{median}}$, and $r_{\text{max}}$ are the predicted scores of the PRM on the corresponding triplet samples. The hyperparameter $\alpha$ is the margin, which defines the degree of separation to impose between the different qualities of samples. The value $\alpha = 0.1$ was chosen for this study. We found empirically that this form of contrastive loss performed better in distinguishing the quality of next-step predictions than losses derived from the Bradley-Terry model of preference \cite{bradleyterry1952} seen commonly in RLHF literature for training reward models given human preference. As such, all PRMs in this study are trained using this contrastive triplet loss.
\subsection*{Test Time Inference}
After having obtained our learned solution operator $S_\theta$, we rely on the stochastic nature of the solution operator to generate a candidate set of predictions $\bigl\{\tilde{u}({t+1})^{(i)}\bigr\}_{i=1}^{B}$ for a given input timestep $\tilde{u}({t})$, where $B$ is referred to as the Branching Factor, which can be tuned during inference. Then, a reward model (ARM or PRM) assigns a score $r_i$ to each pair $\left(\tilde{u}({t}),\tilde{u}({t+1})^i\right)$. The candidate with the highest reward is selected as the successor and fed back into the model to drive the next autoregressive step. This procedure has been highlighted in Algorithm \ref{alg:algorithm}. Since the algorithm involves generating $B$ candidate predictions, the snapshots generated during the autoregressive rollout are specific to the branching factor and hence are identified with $\tilde{u}_B(t) \forall t\in[1,\cdots, T]$. 

\begin{algorithm}[tb]
\caption{Greedy Selection Strategy}
\label{alg:algorithm}
\textbf{Input}: $\tilde{u}(t=1)$ and Branching Factor $B$\\
\textbf{Models}: Base FM $S_\theta$ and Analytical/Process-Reward Model $RM$\\
\textbf{Output}: $\tilde{u}(T)$
\begin{algorithmic}[1] 
\WHILE{$t\leq T$}
\STATE $S_\theta \tilde{u}(t)\to\tilde{u}(t+1)^{(i)} \quad i\in [1,2,\cdots, B] $
\STATE $r_i = RM(\tilde{u}(t),\tilde{u}(t+1)^{(i)}) $
\STATE $\tilde{u}(t+1) = \tilde{u}(t+1)^{i^*}$ where $i^* = \argmax_i r_i $
\STATE $i\to i+1$
\ENDWHILE
\STATE \textbf{return} $\tilde{u}_B(T)$
\end{algorithmic}
\end{algorithm}

\begin{figure*}[h]
    \centering
    \includegraphics[width=0.8\textwidth]{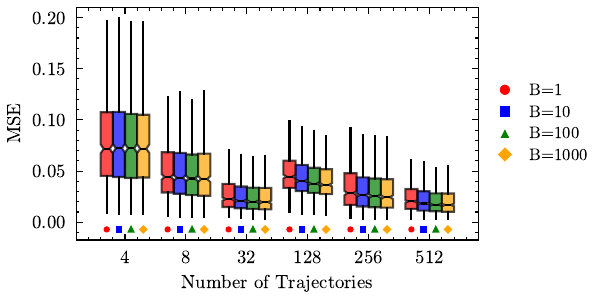}
    \caption{MSE Improvement across various branching factors for ViT-7 models finetuned on different number of RPUI trajectories. As $B$ is increased, MSE of model finetuned on $n_1$ trajectories approaches the MSE of model finetuned on $n_2$ trajectories, where $n_1<n_2$.}
    \label{fig:finetuning_mse}
\end{figure*}

\section*{Results}
\label{sec:results}
All the models have been evaluated on various tasks and various rollout times. Making smarter choices during earlier stages of the autoregressive rollout makes significant improvements in the error observed during the final timestep. 
In addition to the common practice in PDE-ML literature to use MSE as a metric of model accuracy, we also present results on physical conservation laws. Along with evaluating $\text{MSE}(\tilde{u}_B(t),u(t))$, another metric we use to evaluate the per-sample improvement of the rollout is the Sample Gain, 
\begin{align}\label{eq:sample_gain}
    SG_i(t,B) &= \frac{\text{MSE}(\tilde{u}_B(t),u(t))}{\text{MSE}(\tilde{u}_{B=1},(t),u(t))},
\end{align}
where \(u(t=0)=a_i \forall i\in[1, 2, \cdots, N] \notag\).
\subsection*{TTC improves upon unseen examples in pretraining datasets}
Through extensive benchmarks on the test set of pretraining datasets, we show that the greedy selection strategy is able to improve the predictions of the base foundational model. Visualizations of the improvement on the base ViT-3 on the pretraining datasets are shown and quantified in a table in the Supplementary Material.
This improvement is summarized in Table \ref{table:model_comparison}. 
While ARMs require no training, four separate PRMs were trained on the four different pretraining datasets. In other words, the CRP-PRM was trained using the predictions of the base FM on a fraction of the CRP dataset (1024 trajectories used for training CRP-PRM compared to the 8000 CRP trajectories present in the pretraining dataset). To prevent data leakage, PRMs only had access to the training split of the pretraining datasets. The initial conditions present in the test set remained unseen because they were excluded from the training set. With an increase in branching factor during the rollout, the MSE shows monotonic improvement for each timestep, across the RP, CRP, Gauss and KH datasets, while using the PRMs. In addition to improvement in MSE, higher branching factors lead to stronger compliance with mass (Eq \ref{eq:mass_cons}) and energy (Eq \ref{eq:energy_cons}) conservation laws. 
It is worth noting that an analysis of ground truth shows momentum conservation (Eq \ref{eq:momentum_cons}) is violated for many trajectories and this bias shows during inference where the predictions do not necessarily improve momentum conservation with increasing branching factors.

\subsection*{TTC generalizes on unseen out-of-distribution downstream tasks}
TTC produces consistent improvements on unseen out-of-distribution finetuning datasets such as RM and RPUI as shown in visualizations in the Supplementary Material. PRMs for finetuned FMs were trained on the same set of trajectories that were used to train the finetuned FM, e.g., the PRM for RPUI, for the finetuned FM trained on 32 RPUI trajectories, was trained on predictions of the same 32 trajectories. We observe that finetuned FMs trained on different trajectories for a given downstream task improve on their MSE, for increasing branching factors when performing TTC. While improvements on the RPUI dataset are significant (Supplementary Figure \ref{fig:rpui_composite}), MSE improvements on the RM dataset are marginal (Supplementary Figure \ref{fig:RM_composite}) owing to the complex nature of the initial conditions present in the RM dataset, as well as due to the fact that the vanilla finetuned RM models (B=1) perform well out of the box (MSE of 0.035 at timestep T=21). Across the board, PRMs outperform analytical reward models as shown in Supplementary Figure \ref{fig:rpui_composite} and \ref{fig:RM_composite} which alludes to the fact that mass conservation might be violated in the training data as well. However, upon deployment of TTC, the physical conservation quantities show improvement (Supplementary Figure \ref{fig:RPUI_Conservation_composite_Box}, \ref{fig:RM_Conservation_composite_Box}), with increasing branching factor. We also demonstrate that performance of a model finetuned with $n_1$ trajectories approaches the MSE of a model finetuned with $n_2$ trajectories, $n_1<n_2$. While obtaining scaling laws requires further investigation, this observation opens the path forward for achieving data efficiency, e.g., obtaining better performance without any additional training or additional observations/trajectories, as shown in Fig. \ref{fig:finetuning_mse}.
\textbf{Significant Sample Gain across all tasks:} Evaluating the performance of different branching factors by comparing the MSE at a given rollout timestep $t$ leads to a misleading conclusion of a marginal improvement in the average MSE observed over different initial conditions. However, from an operational perspective, the relevant quantity is the percentage gain in performance for a given initial condition, as a function of the branching factor $B$. To this end, we evaluate the sample gain (Eq. \ref{eq:sample_gain}) across different base foundation models and different reward models. Our claims of significant performance improvement while using PRMs are verified by summarizing the sample gain values in Table \ref{table:model_comparison}. The use of analytical reward models often leads to deterioration in performance, which we attribute to the underlying mass conservation violations present in the training dataset.
\section*{Discussion}
Our work introduces a reward-model-driven, test-time computation (TTC) strategy that significantly enhances the accuracy of PDE foundation models, particularly during challenging autoregressive rollouts. Guided by analytical and learned reward models, this method effectively mitigates compounding errors and improves adherence to fundamental physical conservation laws. By extending beyond standard modeling approaches, our work shows that there is considerable promise for “reasoning” in PDE foundation models. Our method differs from conventional predictor-corrector algorithms, due to our use of a problem-agnostic reward signal. Unlike data-specific correctors, which learn to rectify errors particular to a given data distribution, our reward signals, based on universal physical principles such as conservation of mass or energy, are inherently generalizable indicators of solution feasibility. This concept shares close parallels with Physics-Informed Machine Learning (PIML). While PIML traditionally embeds physical constraints directly into the training loss function, our approach instead employs these constraints at inference time. Furthermore, we use the reward signal exclusively to select among candidate solutions, without updating model weights. The TTC framework thus complements the foundation model paradigm by providing an orthogonal pathway to performance improvement, particularly valuable when downstream data for fine-tuning is limited. For practical application of foundation models for PDEs, compute efficiency and training data efficiency requirements are paramount. FM surrogate models for PDEs must necessarily have lower compute cost than high-fidelity simulations, and training data in many critical downstream applications are too expensive to generate beyond a few samples. Therefore, our results show that test-time-compute can address a critical gap in usability of PDE FMs.


Finally, our study encourages broader conceptual and philosophical reflections regarding the meaning of “reasoning” within computational physics and scientific modeling. Unlike in LLMs where correct outputs occupy vast, subjective solution spaces; PDE-governed systems provide objective benchmarks for solution quality. This clarity not only facilitates robust, problem-agnostic reward signals but also prompts deeper consideration of the boundary between inference-time selection and genuine model adaptation. Future research should explore richer forms of adaptive reasoning, potentially integrating limited reinforcement-learning loops, to clarify whether reward-driven TTC simply replicates classical constraint enforcement or represents a genuinely novel inference paradigm.

\section*{Acknowledgements}
Research presented in this article was supported by the Laboratory Directed Research and Development program of Los Alamos National Laboratory under project numbers 20250637DI and 20250639DI. 
Los Alamos National Laboratory is operated by Triad National Security, LLC, for the National Nuclear Security Administration of U.S. Department of Energy (Contract No. 89233218CNA000001). 
\bibliographystyle{naturemag}
\bibliography{arxivref}

\include{Supplementary}
\end{document}

%% file: Supplementary.tex


\title{Supplementary Material for: \\ Towards Reasoning for PDE Foundation Models: A Reward-Model-Driven Inference-Time-Scaling Algorithm}

\renewcommand{\thefigure}{S\arabic{figure}}
\renewcommand{\thetable}{S\arabic{table}}
\setcounter{figure}{0}
\setcounter{table}{0}
\tableofcontents
\newpage
\section{Hyperparameters for Pre-training and Finetuning Models}
\begin{table}[ht]
\centering
\begin{tabular}{ll}
\toprule
\textbf{Hyperparameter} & \textbf{Value} \\
\midrule
Image size & 128$\times$128 \\
Patch size & [3,5,7] \\
Input channels & 5 \\
Embedding dimension & 256 \\
Transformer depth & 6 \\
Number of heads & 8 \\
MLP ratio & 4 \\
Dropout rate & 0.1 \\
Learning rate & $5 \times 10^{-6}$ \\
Weight decay & $1 \times 10^{-7}$ \\
\midrule
Finetune learning rate & $1 \times 10^{-5}$ \\
Finetune optimizer & AdamW \\
Finetune weight decay & 0.01 \\
\bottomrule
\end{tabular}
\caption{Hyperparameter settings used for the Vision Transformer image-to-image translation model.}
\label{tab:hyperparams}
\end{table}
\section{Computational Resources} \label{sec:comp_resources}
All pretrainings were performed in parallel on  a GPU node comprised of 8 NVIDIA H100 GPUs, each with 80GB VRAM, 1 TB DDR4 RAM and 2 Intel Xeon 8470 processors. Pretraining each model on the four Compressible Euler datasets took 16 hours to train (approximately 40 epochs). Training of the process-reward models required generating training data which comprised of predictions of the pretrained model over the training set. Once the PRM training data was generated, training the PRM (same architecture as the base foundational model) required 16 hours. After training the PRM, generating predictions over 500 trajectories of any dataset with branching factor 1000 required 2 hours (worst time estimate) while using all 8 GPUs. 
\newpage
\section{Pretrain Visualizations on Test Set}
\begin{figure}[H]
    \centering
    \includegraphics[width=1\textwidth]{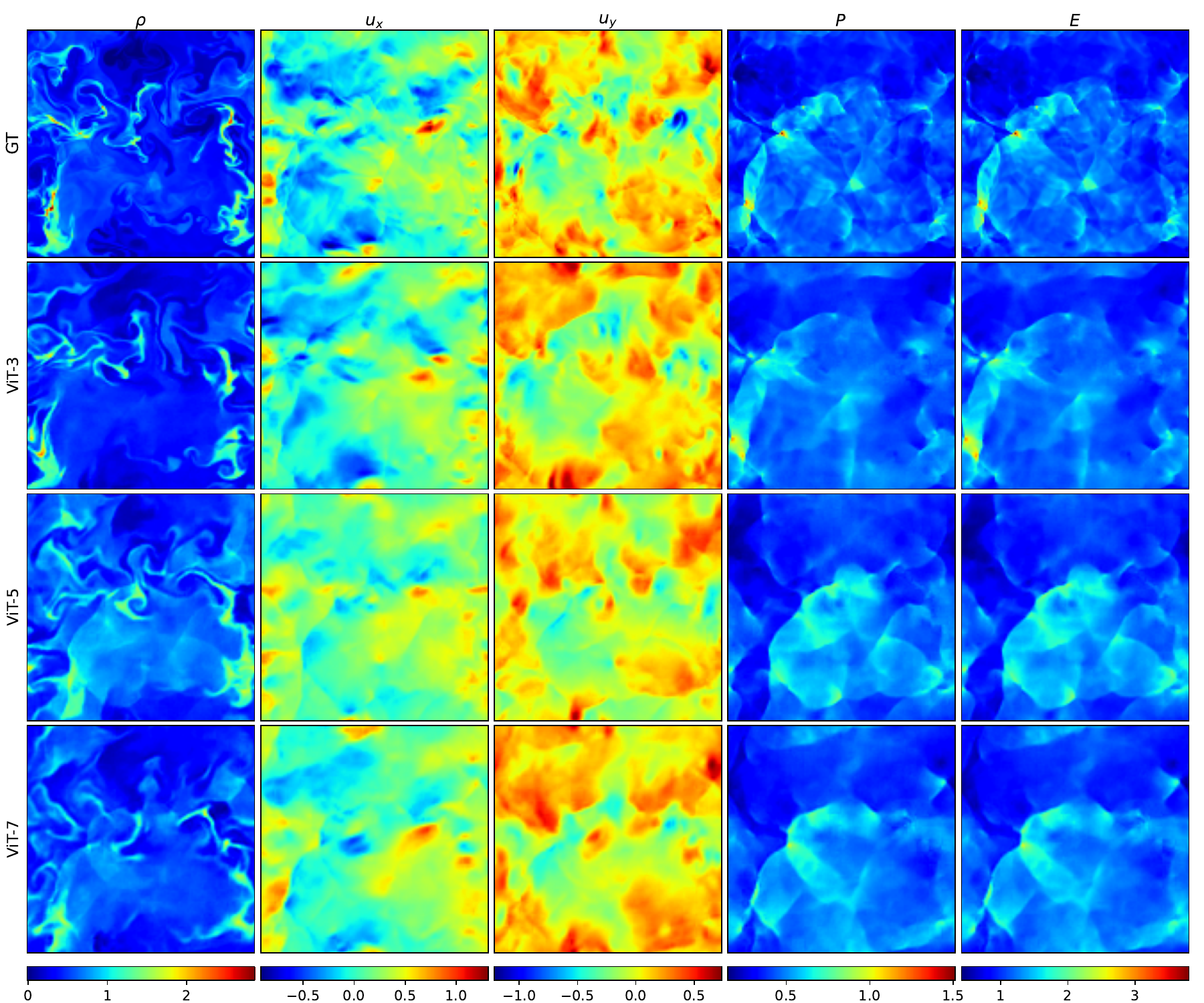}
    \caption{Rollout Performance of models on CRP}
    \label{fig:crp_prm_rollout}
\end{figure}

\begin{figure}[H]
    \centering
    \includegraphics[width=1\textwidth]{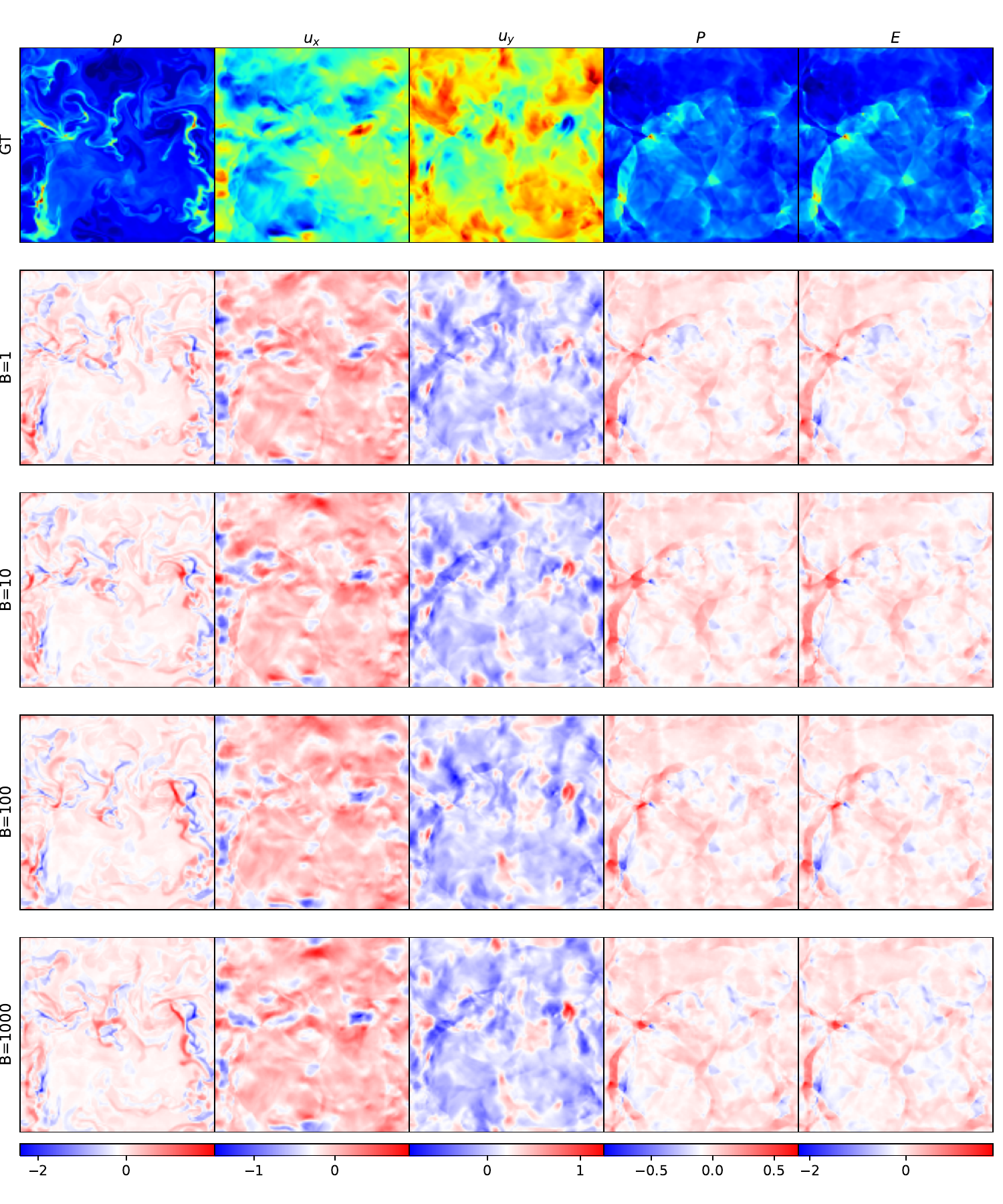}
    \caption{Comparison of PRM on CRP with ViT-3}
    \label{fig:crp_prm_diff}
\end{figure}

\begin{figure}[H]
    \centering
    \includegraphics[width=1\textwidth]{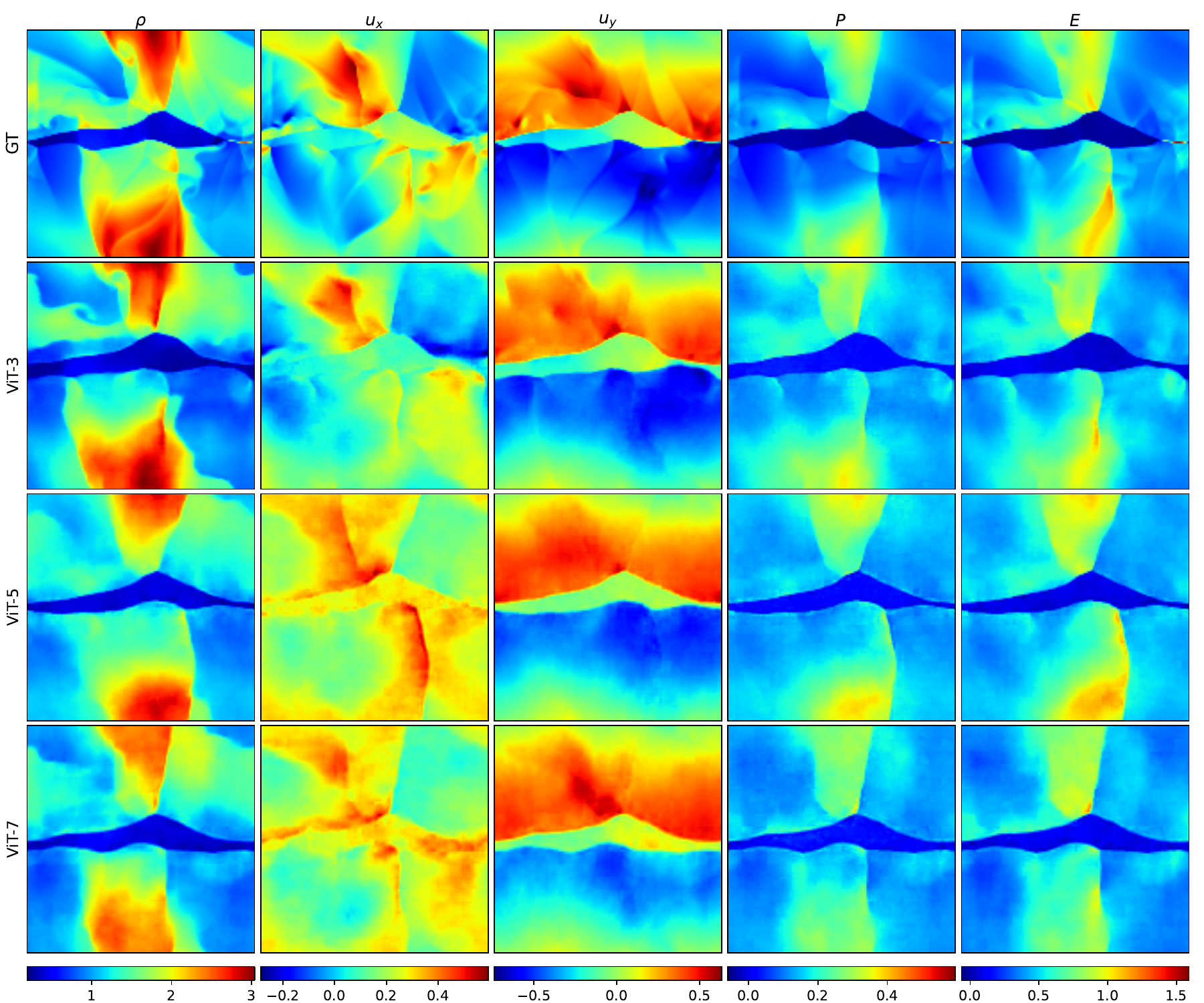}
    \caption{Rollout Performance of models on RP}
    \label{fig:rp_prm_rollout}
\end{figure}

\begin{figure}[H]
    \centering
    \includegraphics[width=1\textwidth]{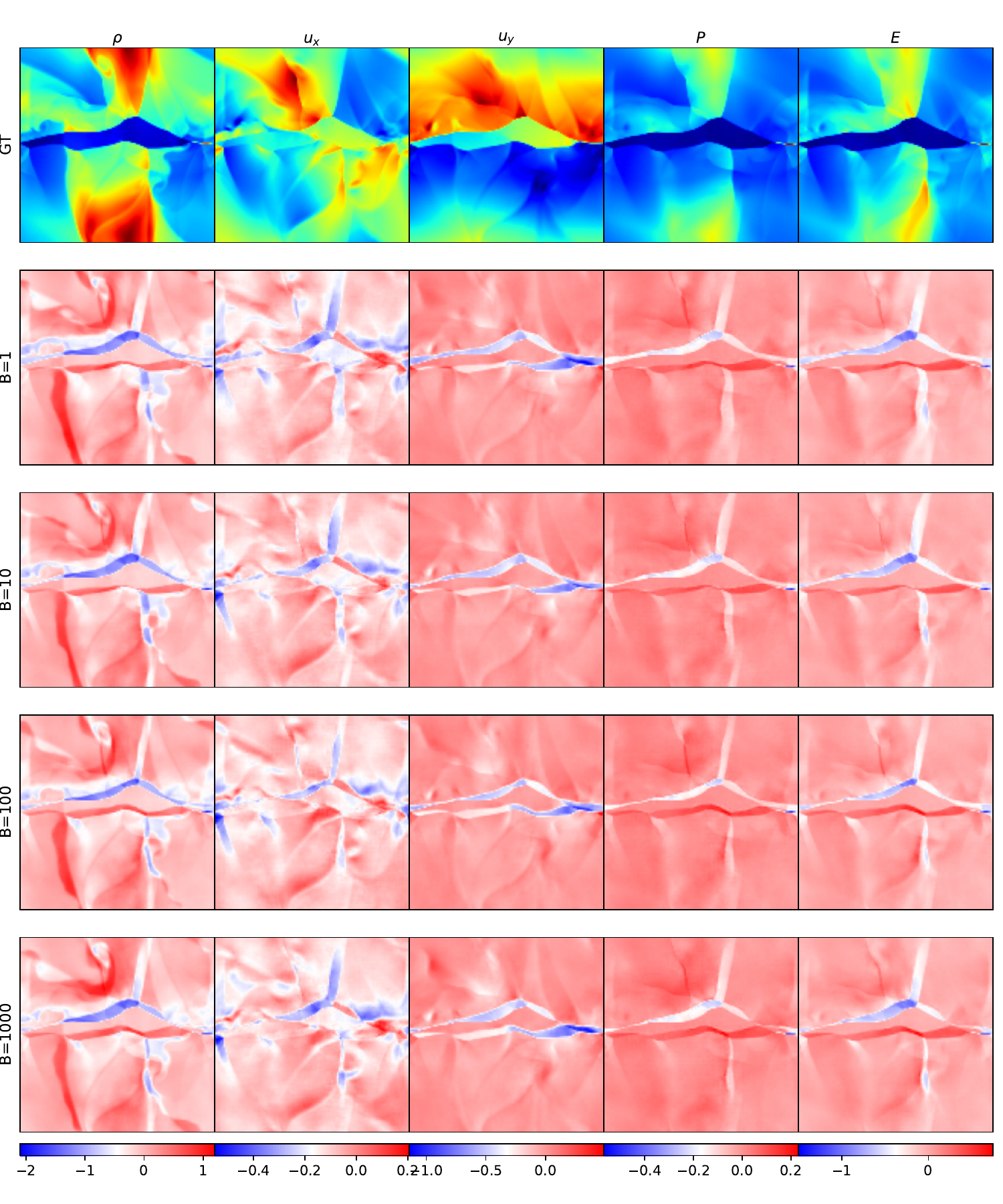}
    \caption{Comparison of PRM on RP with ViT-3}
    \label{fig:rp_prm_diff}
\end{figure}

\begin{figure}[H]
    \centering
    \includegraphics[width=1\textwidth]{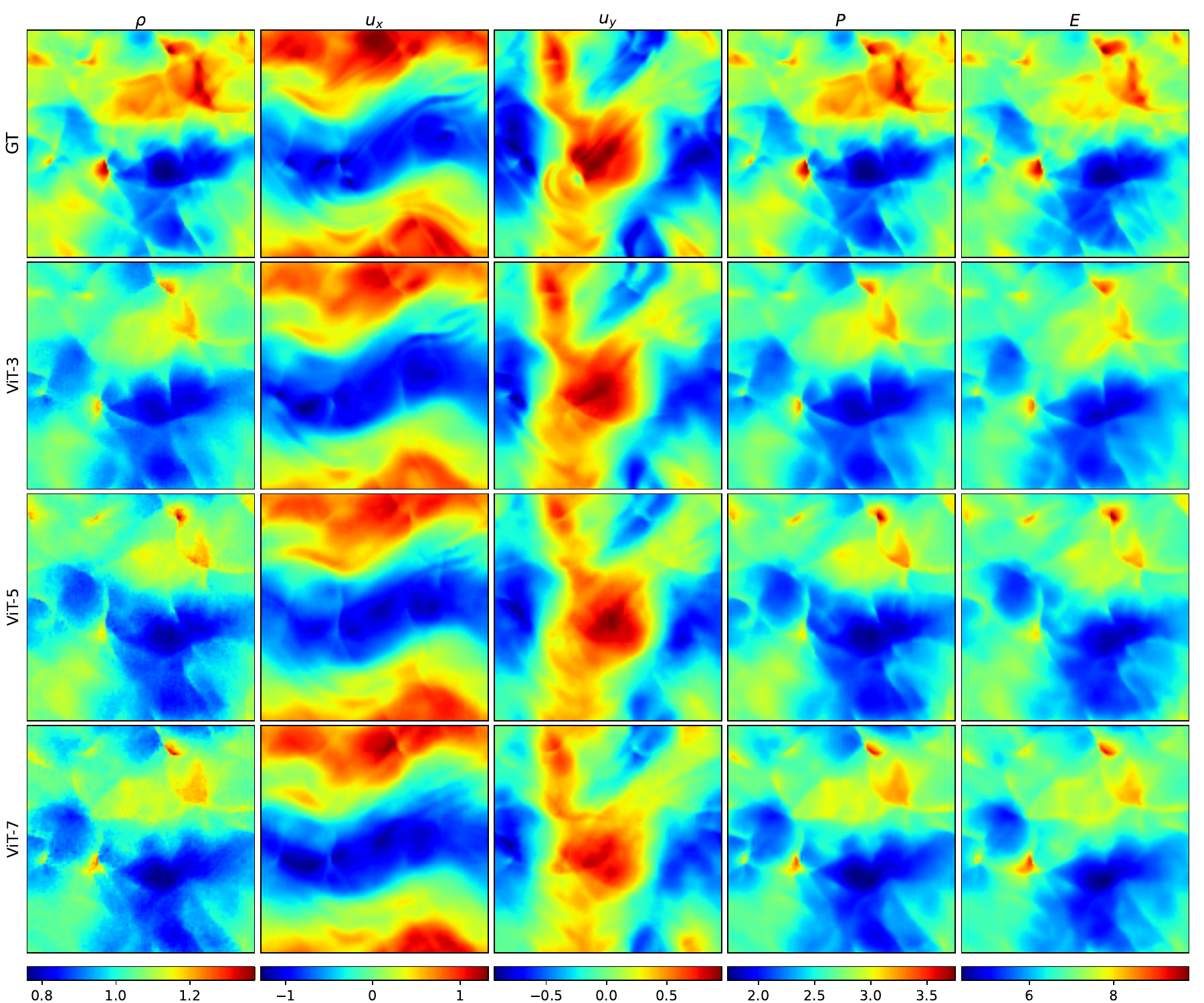}
    \caption{Rollout Performance of models on RP}
    \label{fig:gauss_prm_rollout}
\end{figure}

\begin{figure}[H]
    \centering
    \includegraphics[width=1\textwidth]{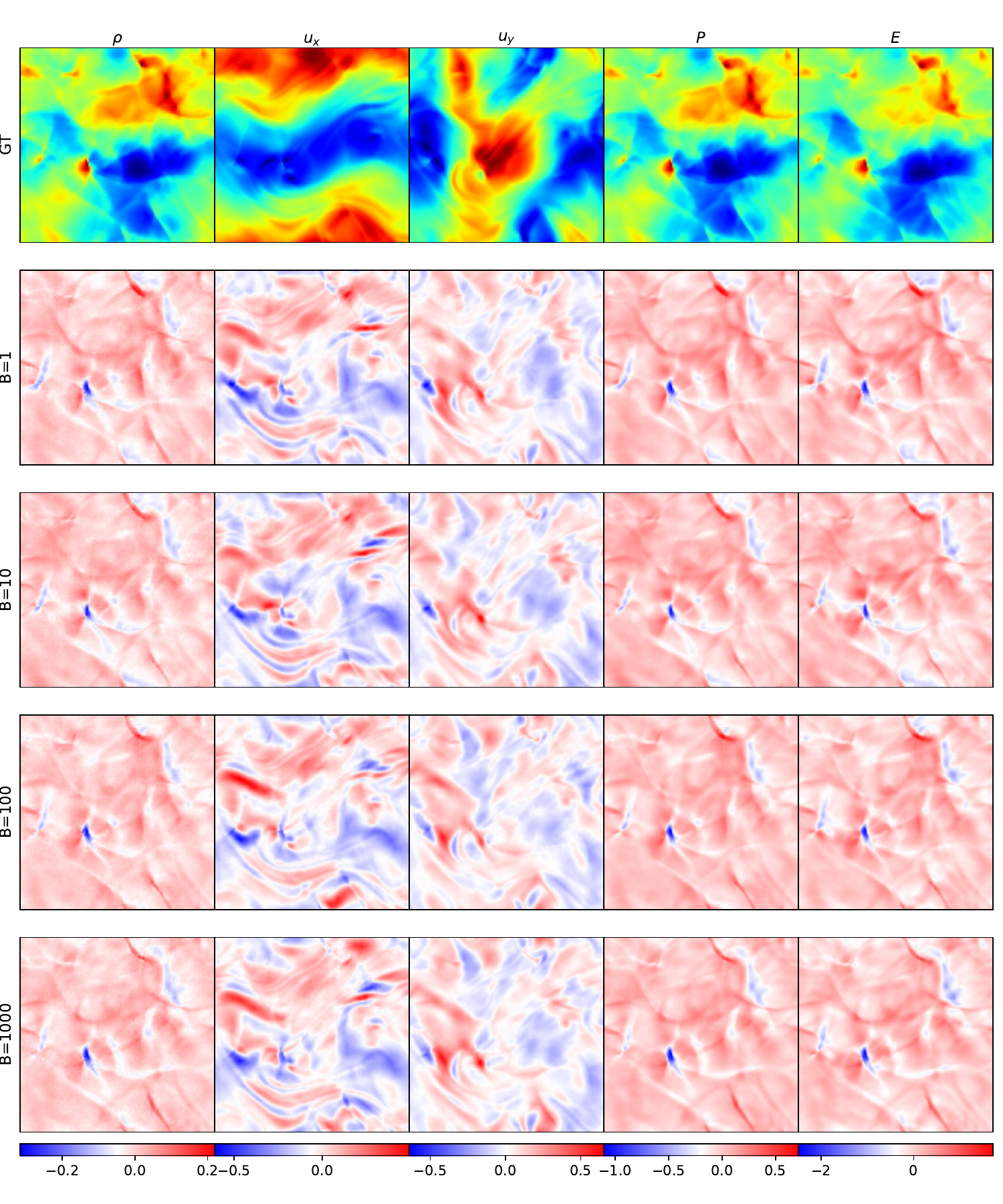}
    \caption{Comparison of PRM on RP with ViT-3}
    \label{fig:gauss_prm_diff}
\end{figure}

\begin{figure}[H]
    \centering
    \includegraphics[width=1\textwidth]{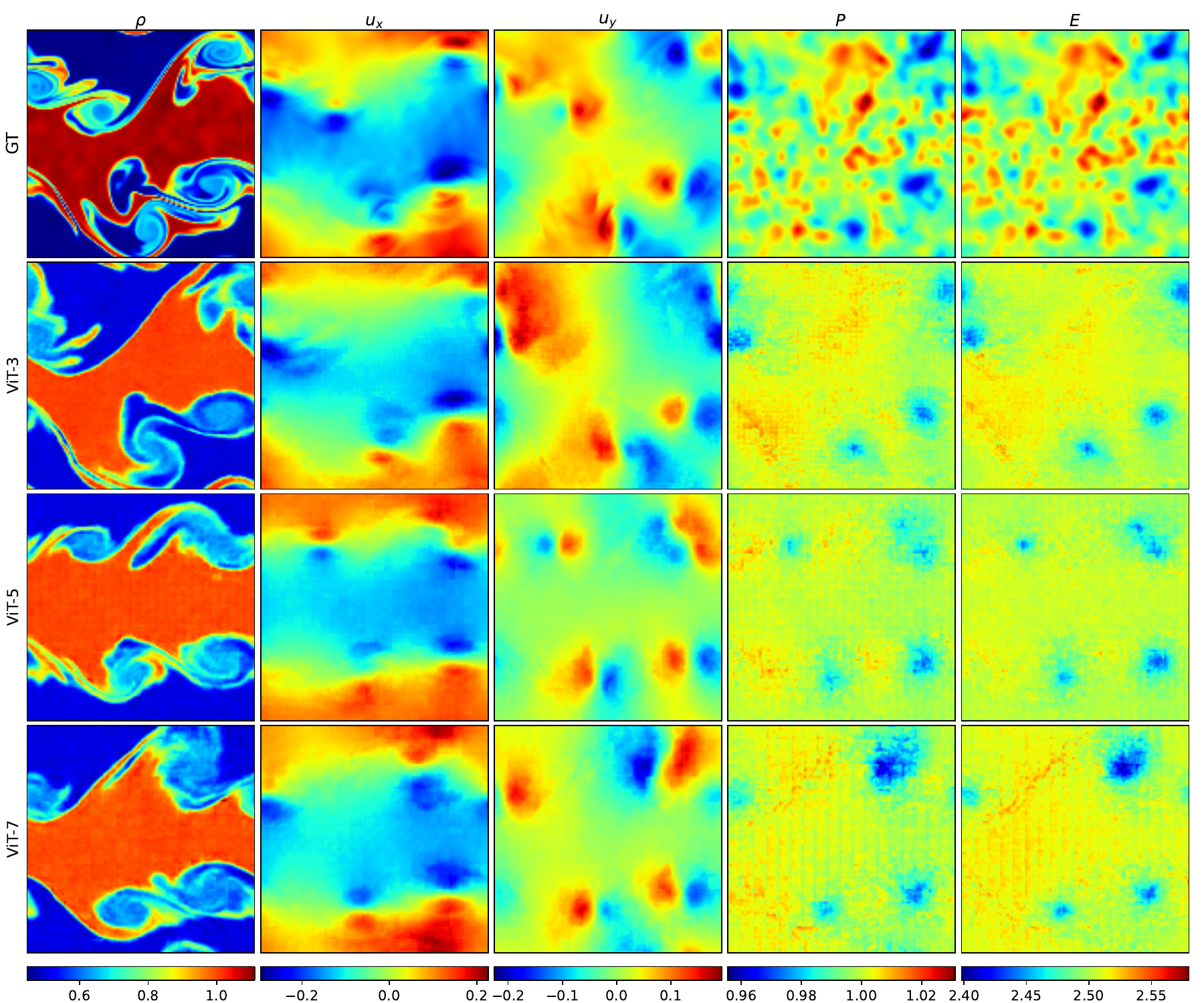}
    \caption{Rollout Performance of models on KH}
    \label{fig:kh_prm_rollout}
\end{figure}

\begin{figure}[H]
    \centering
    \includegraphics[width=1\textwidth]{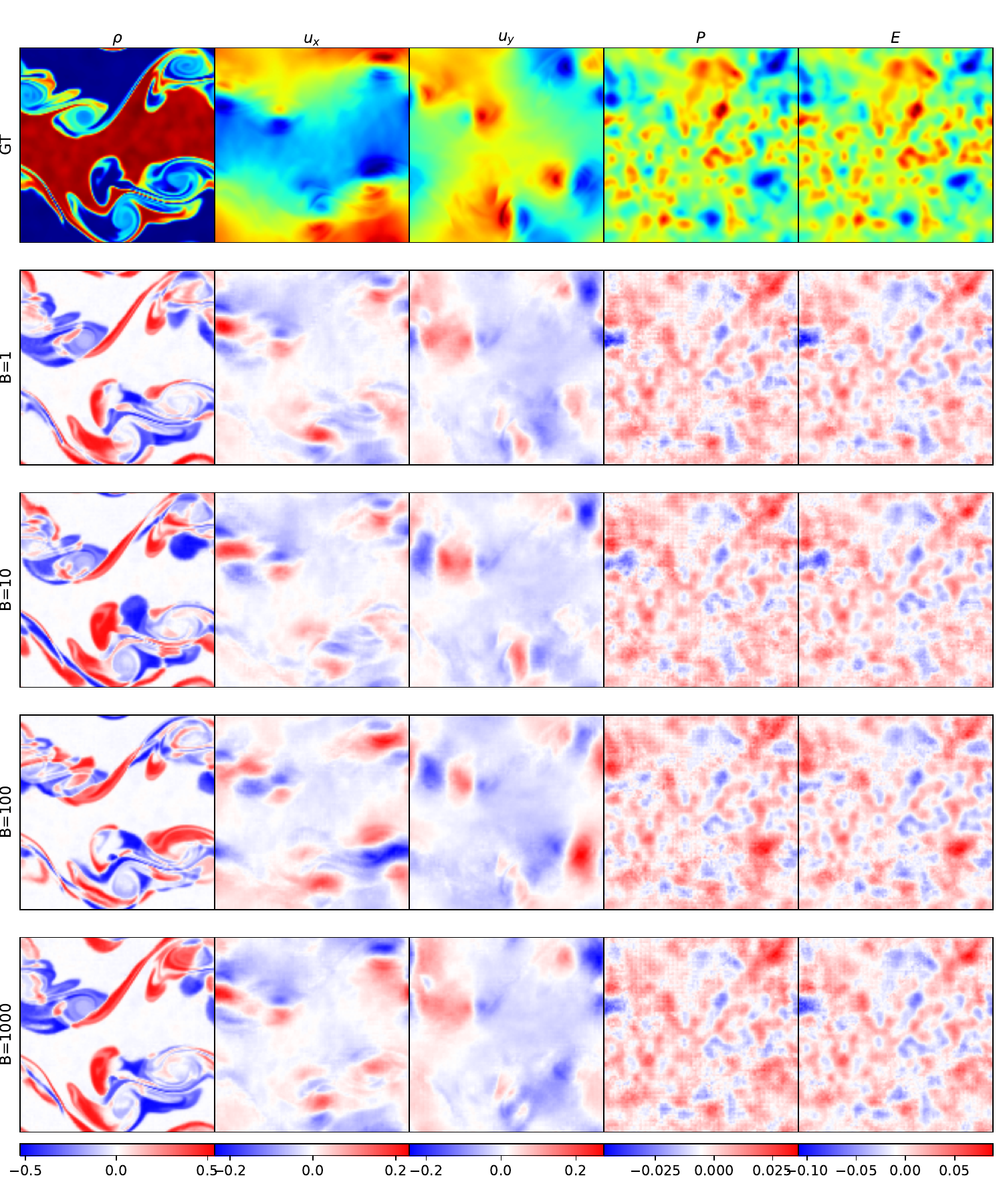}
    \caption{Comparison of PRM on KH with ViT-3}
    \label{fig:kh_prm_diff}
\end{figure}

\section{Finetuning Visualizations on Downstream Tasks}
\begin{figure}[H]
    \centering
    \includegraphics[width=1\textwidth]{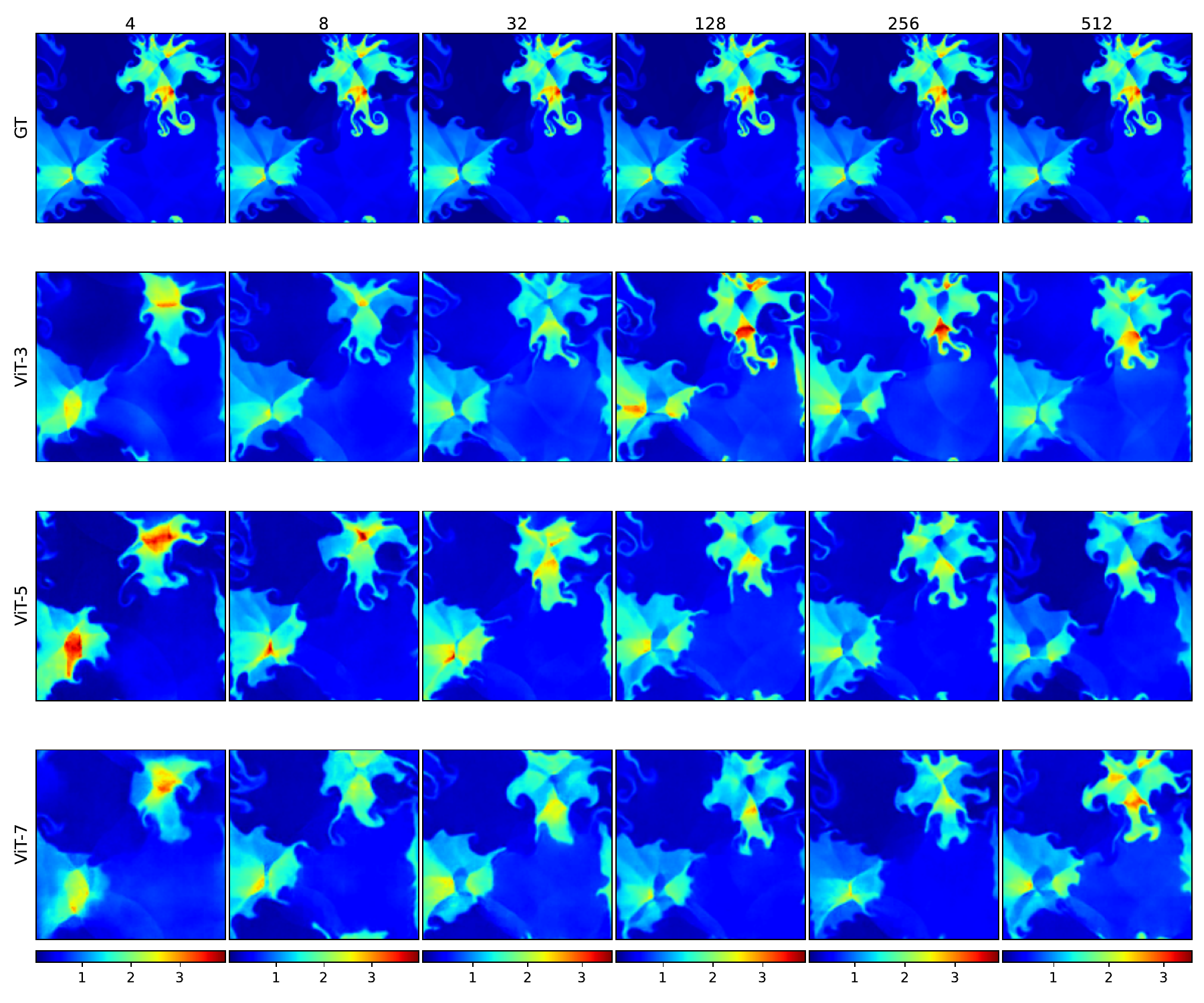}
    \caption{Rollout Performance of models on RPUI}
    \label{fig:rpui_prm_rollout}
\end{figure}

\begin{figure}[H]
    \centering
    \includegraphics[width=1\textwidth]{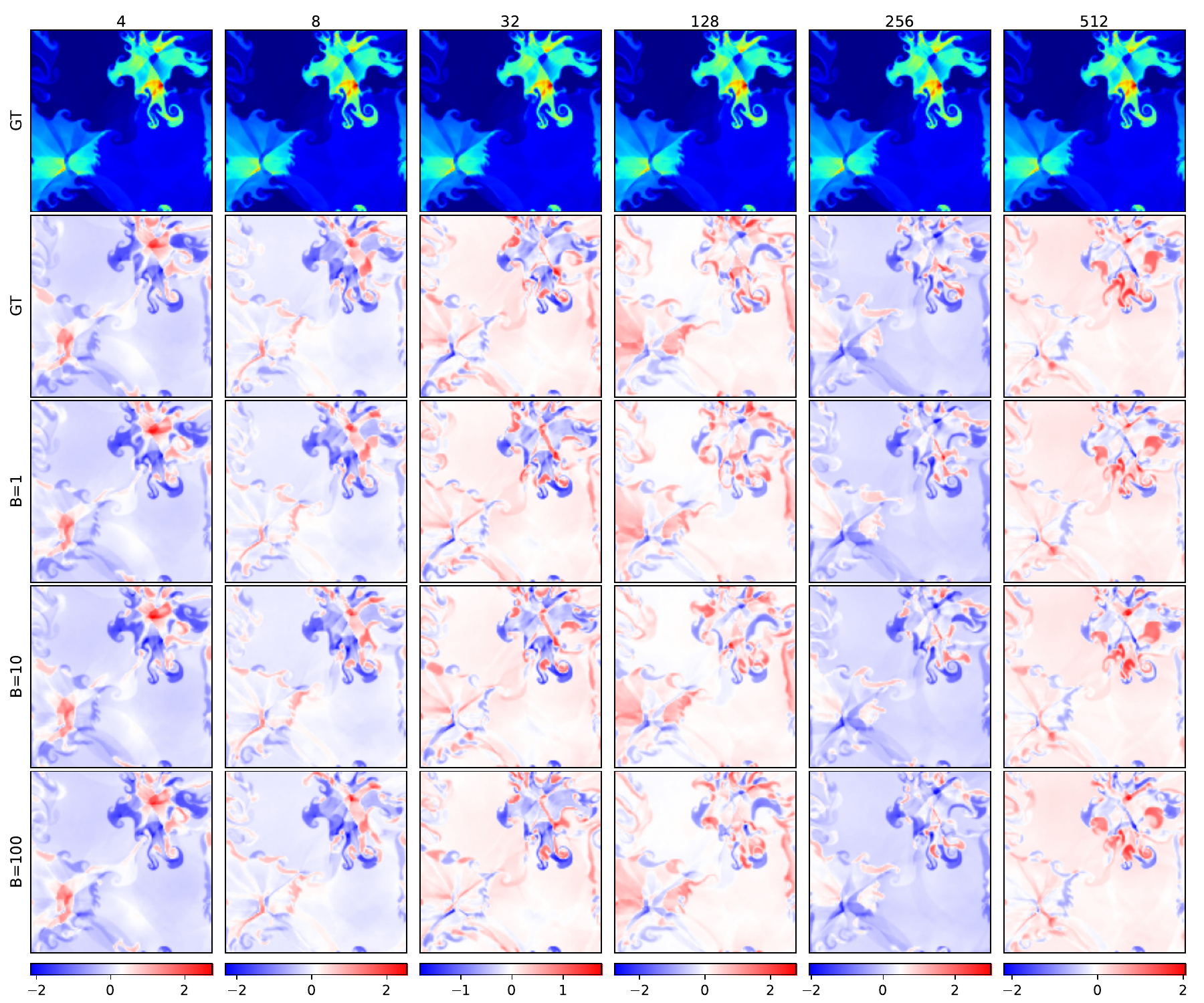}
    \caption{Comparison of PRM on RPUI with ViT-3}
    \label{fig:rpui_prm_diff}
\end{figure}

\begin{figure}[H]
    \centering
    \includegraphics[width=1\textwidth]{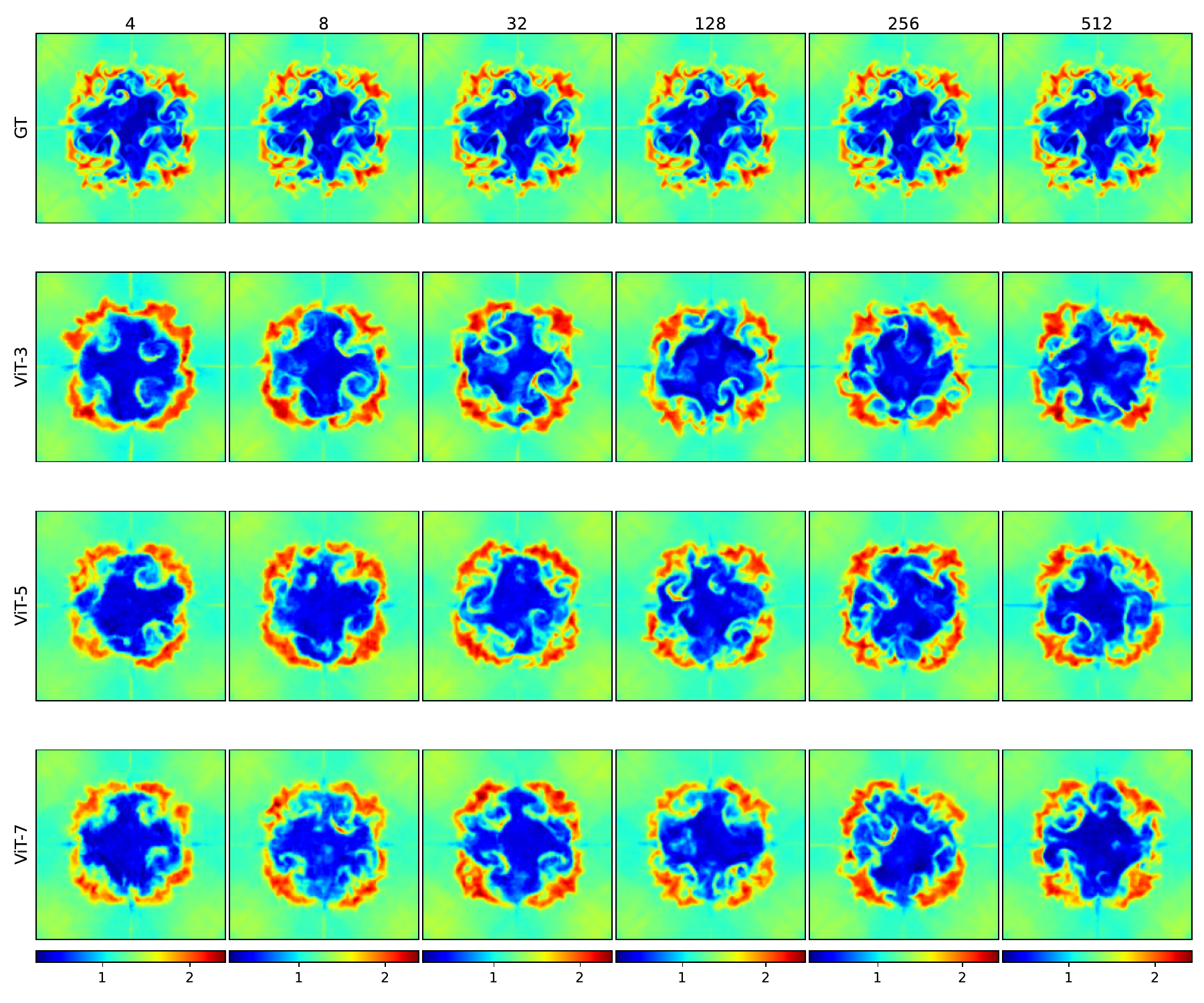}
    \caption{Rollout Performance of models on RM}
    \label{fig:rm_prm_rollout}
\end{figure}

\begin{figure}[H]
    \centering
    \includegraphics[width=1\textwidth]{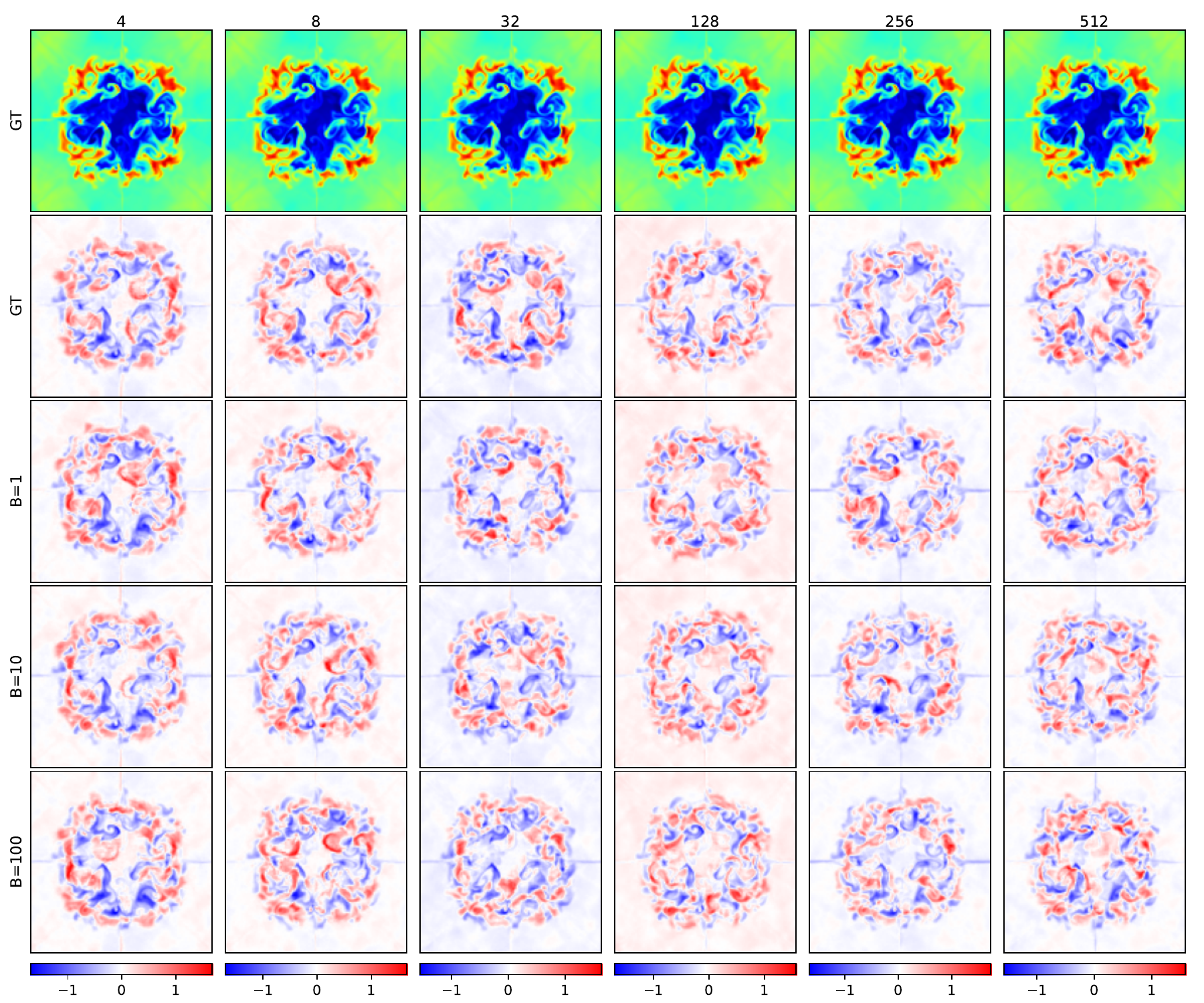}
    \caption{Comparison of PRM on RM with ViT-3}
    \label{fig:rm_prm_diff}
\end{figure}

\section{Pretraining TTC Measurements}

\begin{figure}[H]
    \begin{subfigure}[t]{0.5\textwidth}
        \includegraphics[width=\textwidth]{pdf_files/Pretrain_prm/CRP_10.pdf}
    \caption{}
        \label{fig:crp_prm}
    \end{subfigure}
    \hfill
    \begin{subfigure}[t]{0.5\textwidth}
        \includegraphics[width=\textwidth]{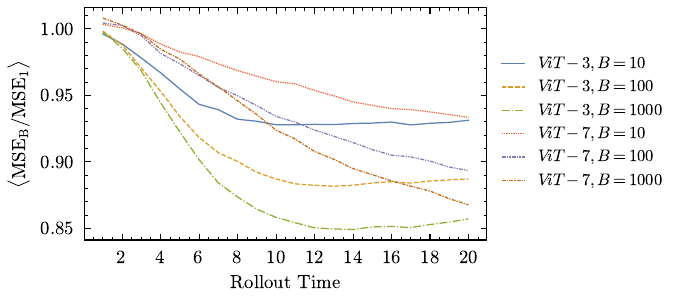}
    \caption{}
        \label{fig:crp_prm_ratio}
    \end{subfigure}
    \hfill
    \begin{subfigure}[t]{0.5\textwidth}
        \includegraphics[width=\textwidth]{pdf_files/Pretrain_Mass/CRP_10.pdf}
    \caption{}
        \label{fig:crp_mass}
    \end{subfigure}
    \hfill
    \begin{subfigure}[t]{0.5\textwidth}
        \includegraphics[width=\textwidth]{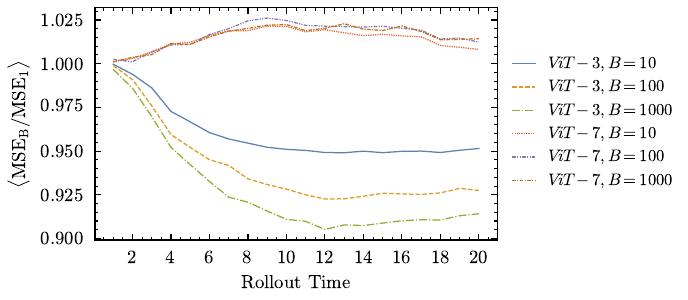}
    \caption{}
        \label{fig:crp_mass_ratio}
    \end{subfigure}
    
    \caption{Performance on CRP for PRM (a,b) and Mass Conservation (c,d) reward model}
    \label{fig:crp_composite}
\end{figure}

\begin{figure}[H]
    \begin{subfigure}[t]{0.5\textwidth}
        \includegraphics[width=\textwidth]{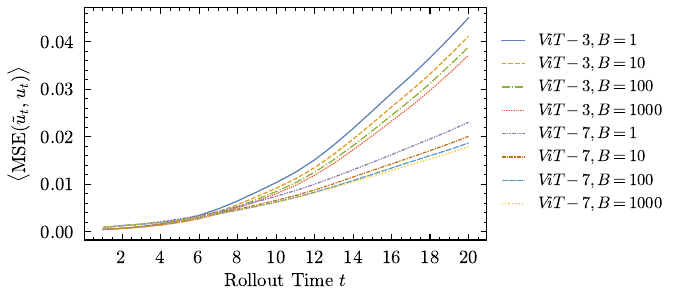}
    \caption{}
        \label{fig:rp_prm}
    \end{subfigure}
    \hfill
    \begin{subfigure}[t]{0.5\textwidth}
        \includegraphics[width=\textwidth]{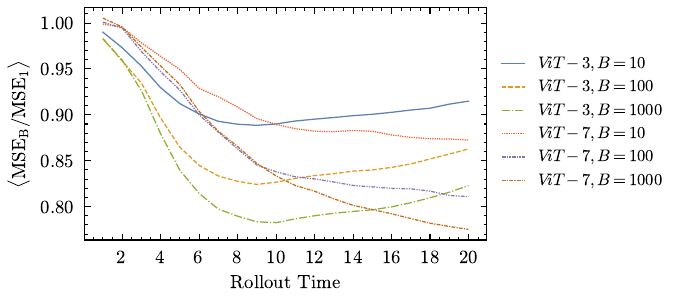}
    \caption{}
        \label{fig:rp_prm_ratio}
    \end{subfigure}
    \hfill
    \begin{subfigure}[t]{0.5\textwidth}
        \includegraphics[width=\textwidth]{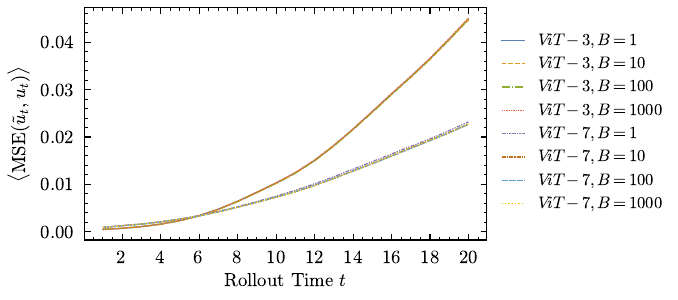}
    \caption{}
        \label{fig:rp_mass}
    \end{subfigure}
    \hfill
    \begin{subfigure}[t]{0.5\textwidth}
        \includegraphics[width=\textwidth]{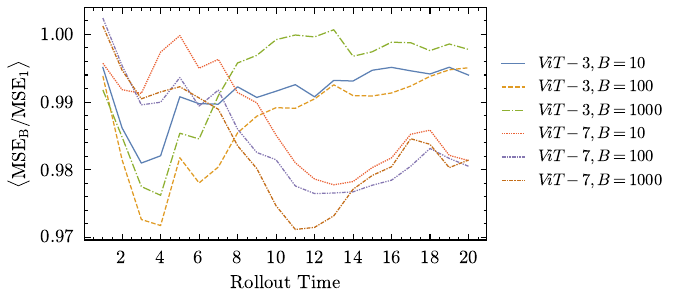}
    \caption{}
        \label{fig:rp_mass_ratio}
    \end{subfigure}
    
    \caption{Performance on RP for PRM (a,b) and Mass Conservation (c,d) reward model}
    \label{fig:rp_composite}
\end{figure}

\begin{figure}[H]
    \begin{subfigure}[t]{0.5\textwidth}
        \includegraphics[width=\textwidth]{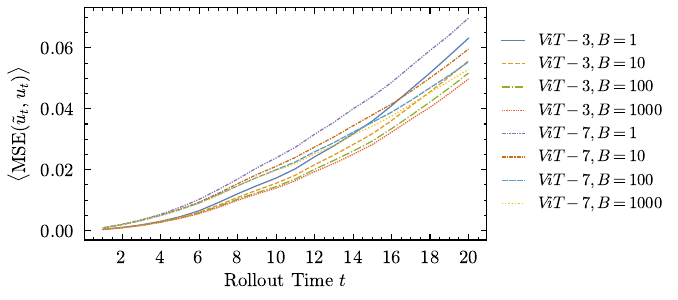}
    \caption{}
        \label{fig:gauss_prm}
    \end{subfigure}
    \hfill
    \begin{subfigure}[t]{0.5\textwidth}
        \includegraphics[width=\textwidth]{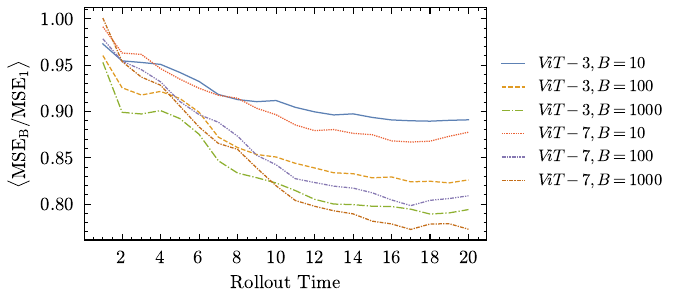}
    \caption{}
        \label{fig:Gauss_prm_ratio}
    \end{subfigure}
    \hfill
    \begin{subfigure}[t]{0.5\textwidth}
        \includegraphics[width=\textwidth]{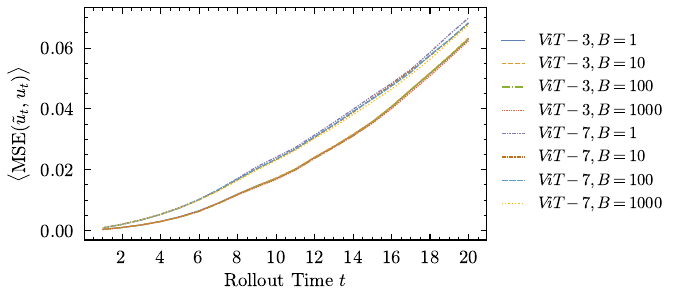}
    \caption{}
        \label{fig:Gauss_mass}
    \end{subfigure}
    \hfill
    \begin{subfigure}[t]{0.5\textwidth}
        \includegraphics[width=\textwidth]{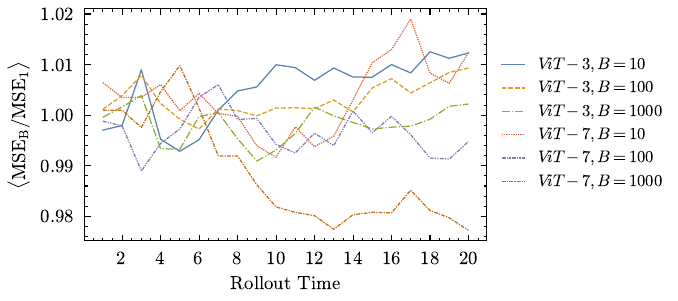}
    \caption{}
        \label{fig:Gauss_mass_ratio}
    \end{subfigure}
    
    \caption{Performance on Gauss for PRM (a,b) and Mass Conservation (c,d) reward model}
    \label{fig:Gauss_composite}
\end{figure}

\begin{figure}[H]
    \begin{subfigure}[t]{0.5\textwidth}
        \includegraphics[width=\textwidth]{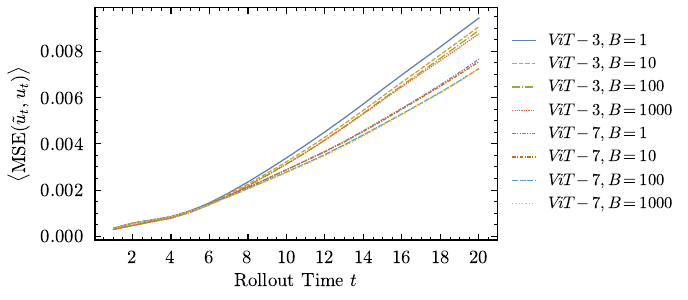}
    \caption{}
        \label{fig:KH_prm}
    \end{subfigure}
    \hfill
    \begin{subfigure}[t]{0.5\textwidth}
        \includegraphics[width=\textwidth]{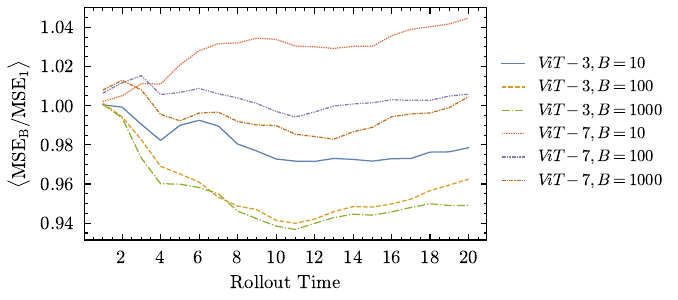}
    \caption{}
        \label{fig:KH_prm_ratio}
    \end{subfigure}
    \hfill
    \begin{subfigure}[t]{0.5\textwidth}
        \includegraphics[width=\textwidth]{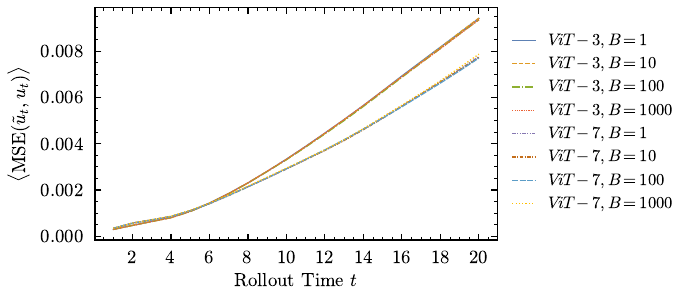}
    \caption{}
        \label{fig:KH_mass}
    \end{subfigure}
    \hfill
    \begin{subfigure}[t]{0.5\textwidth}
        \includegraphics[width=\textwidth]{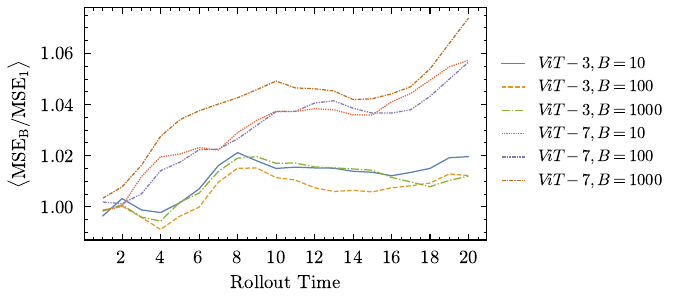}
    \caption{}
        \label{fig:KH_mass_ratio}
    \end{subfigure}
    
    \caption{Performance on KH for PRM (a,b) and Mass Conservation (c,d) reward model}
    \label{fig:kh_composite}
\end{figure}
\section{Pretraining Physical Metrics Measurements}
\begin{figure}[H]
    \begin{subfigure}[t]{0.5\textwidth}
        \includegraphics[width=\textwidth]{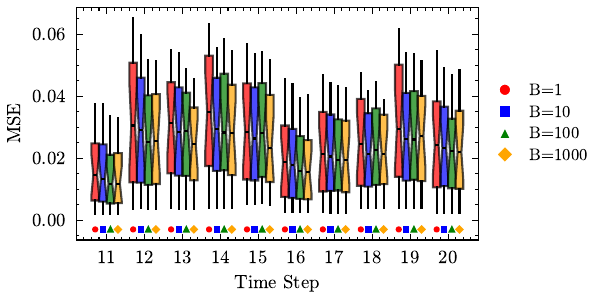}
    \caption{}
        \label{fig:CRP_prm_MSE_Box}
    \end{subfigure}
    \hfill
    \begin{subfigure}[t]{0.5\textwidth}
        \includegraphics[width=\textwidth]{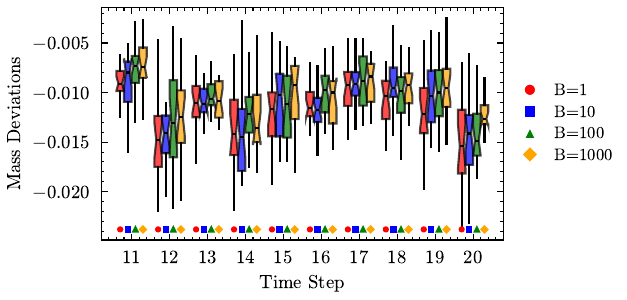}
    \caption{}
        \label{fig:CRP_prm_Mass_Box}
    \end{subfigure}
    \hfill
    \begin{subfigure}[t]{0.5\textwidth}
        \includegraphics[width=\textwidth]{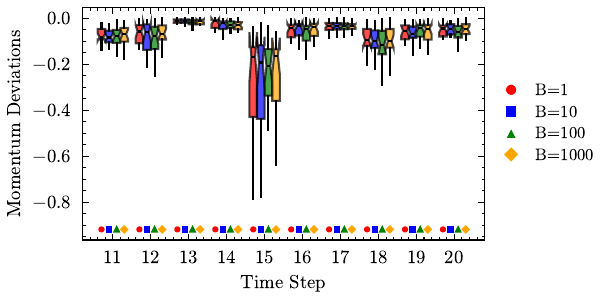}
    \caption{}
        \label{fig:CRP_prm_Momentum_Box}
    \end{subfigure}
    \hfill
    \begin{subfigure}[t]{0.5\textwidth}
        \includegraphics[width=\textwidth]{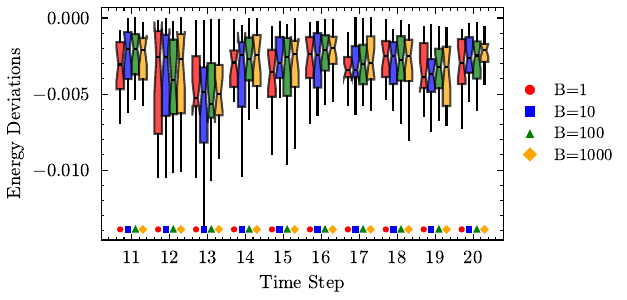}
    \caption{}
        \label{fig:CRP_prm_Energy_Box}
    \end{subfigure}
    
    \caption{Conservation Metrics on CRP for PRM Reward model}
    \label{fig:CRP_Conservation_composite_Box}
\end{figure}

\begin{figure}[H]
    \begin{subfigure}[t]{0.5\textwidth}
        \includegraphics[width=\textwidth]{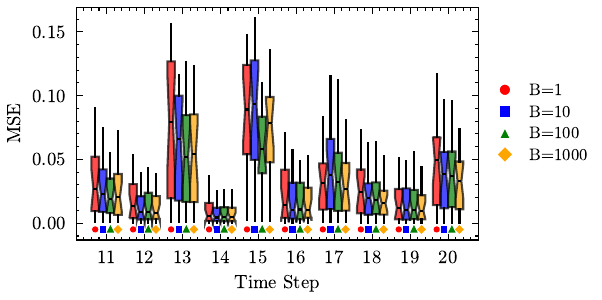}
    \caption{}
        \label{fig:Gauss_prm_MSE_Box}
    \end{subfigure}
    \hfill
    \begin{subfigure}[t]{0.5\textwidth}
        \includegraphics[width=\textwidth]{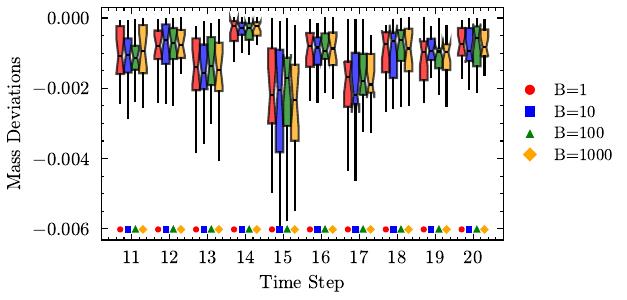}
    \caption{}
        \label{fig:Gauss_prm_Mass_Box}
    \end{subfigure}
    \hfill
    \begin{subfigure}[t]{0.5\textwidth}
        \includegraphics[width=\textwidth]{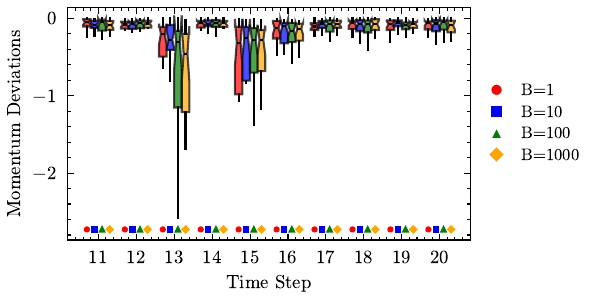}
    \caption{}
        \label{fig:Gauss_prm_Momentum_Box}
    \end{subfigure}
    \hfill
    \begin{subfigure}[t]{0.5\textwidth}
        \includegraphics[width=\textwidth]{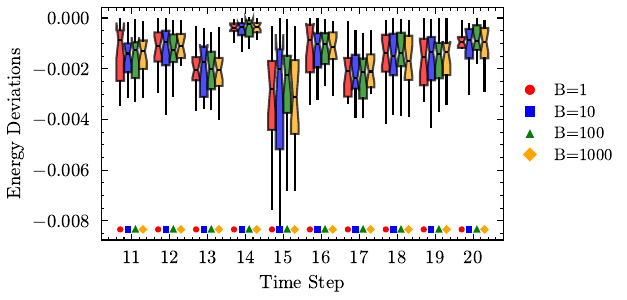}
    \caption{}
        \label{fig:Gauss_prm_Energy_Box}
    \end{subfigure}
    
    \caption{Conservation Metrics on Gauss for PRM Reward model}
    \label{fig:Gauss_Conservation_composite_Box}
\end{figure}

\begin{figure}[H]
    \begin{subfigure}[t]{0.5\textwidth}
        \includegraphics[width=\textwidth]{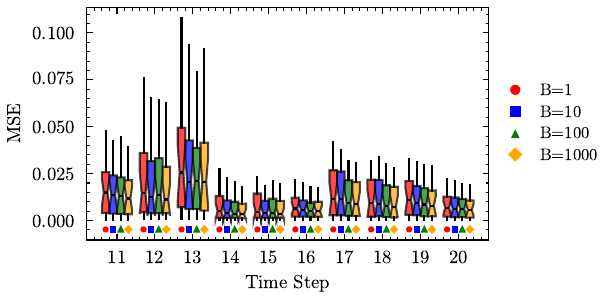}
    \caption{}
        \label{fig:RP_prm_MSE_Box}
    \end{subfigure}
    \hfill
    \begin{subfigure}[t]{0.5\textwidth}
        \includegraphics[width=\textwidth]{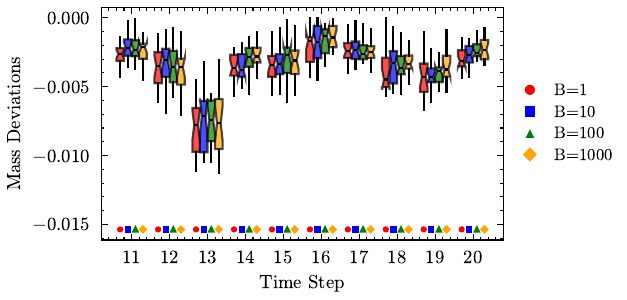}
    \caption{}
        \label{fig:RP_prm_Mass_Box}
    \end{subfigure}
    \hfill
    \begin{subfigure}[t]{0.5\textwidth}
        \includegraphics[width=\textwidth]{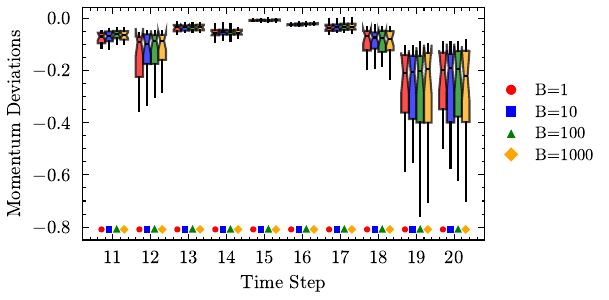}
    \caption{}
        \label{fig:RP_prm_Momentum_Box}
    \end{subfigure}
    \hfill
    \begin{subfigure}[t]{0.5\textwidth}
        \includegraphics[width=\textwidth]{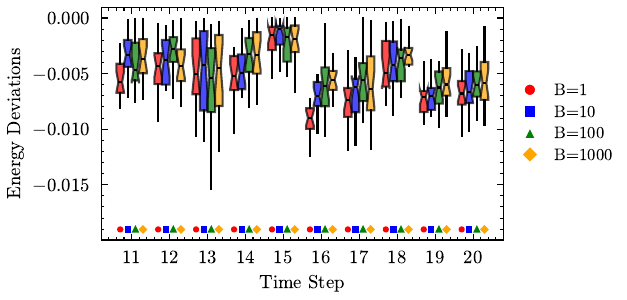}
    \caption{}
        \label{fig:RP_prm_Energy_Box}
    \end{subfigure}
    
    \caption{Conservation Metrics on RP for PRM Reward model}
    \label{fig:RP_Conservation_composite_Box}
\end{figure}

\begin{figure}[H]
    \begin{subfigure}[t]{0.5\textwidth}
        \includegraphics[width=\textwidth]{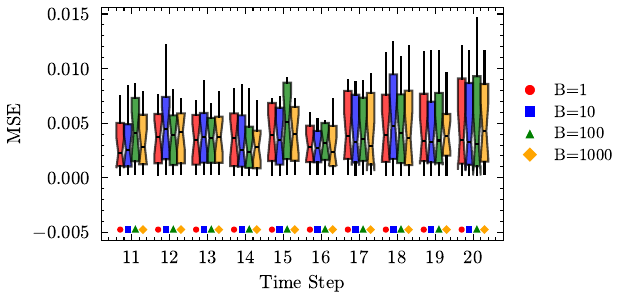}
    \caption{}
        \label{fig:KH_prm_MSE_Box}
    \end{subfigure}
    \hfill
    \begin{subfigure}[t]{0.5\textwidth}
        \includegraphics[width=\textwidth]{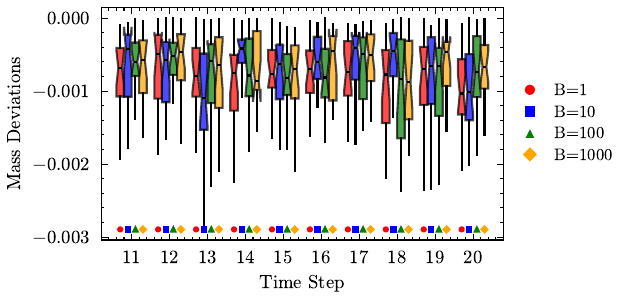}
    \caption{}
        \label{fig:KH_prm_Mass_Box}
    \end{subfigure}
    \hfill
    \begin{subfigure}[t]{0.5\textwidth}
        \includegraphics[width=\textwidth]{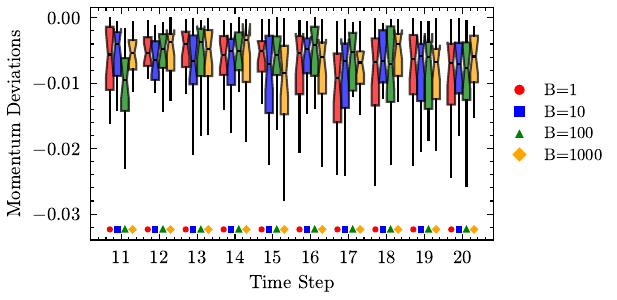}
    \caption{}
        \label{fig:KH_prm_Momentum_Box}
    \end{subfigure}
    \hfill
    \begin{subfigure}[t]{0.5\textwidth}
        \includegraphics[width=\textwidth]{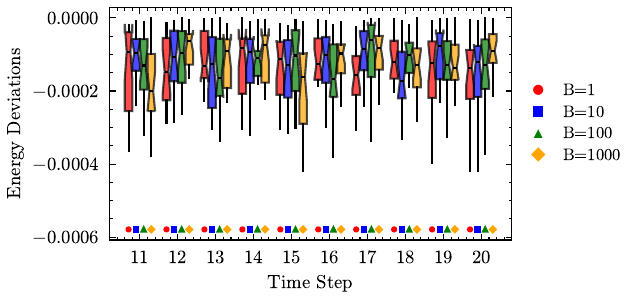}
    \caption{}
        \label{fig:KH_prm_Energy_Box}
    \end{subfigure}
    
    \caption{Conservation Metrics on KH for PRM Reward model}
    \label{fig:KH_Conservation_composite_Box}
\end{figure}

\section{Finetuning TTC Measurements}

\begin{figure}[H]
    \begin{subfigure}[t]{0.5\textwidth}
        \includegraphics[width=\textwidth]{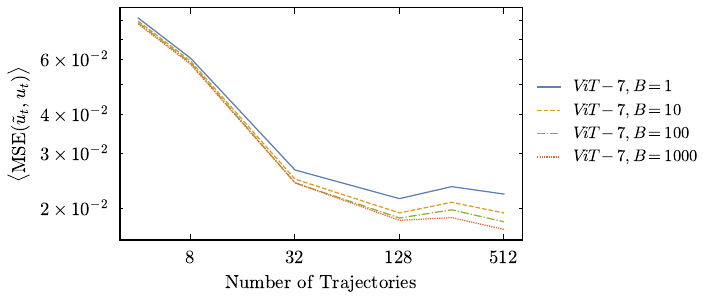}
    \caption{}
        \label{fig:rpui_prm}
    \end{subfigure}
    \hfill
    \begin{subfigure}[t]{0.5\textwidth}
        \includegraphics[width=\textwidth]{pdf_files/Finetune_prm/Ratio_RPUI_10.pdf}
    \caption{}
        \label{fig:rpui_prm_ratio}
    \end{subfigure}
    \hfill
    \begin{subfigure}[t]{0.5\textwidth}
        \includegraphics[width=\textwidth]{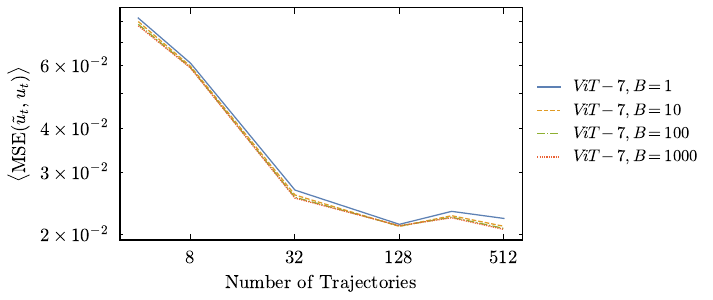}
    \caption{}
        \label{fig:rpui_mass}
    \end{subfigure}
    \hfill
    \begin{subfigure}[t]{0.5\textwidth}
        \includegraphics[width=\textwidth]{pdf_files/Finetune_Mass/Ratio_RPUI_10.pdf}
    \caption{}
        \label{fig:rpui_mass_ratio}
    \end{subfigure}
    
    \caption{Performance on RPUI for PRM (a,b) and Mass Conservation (c,d) reward model}
    \label{fig:rpui_composite}
\end{figure}

\begin{figure}[H]
    \begin{subfigure}[t]{0.5\textwidth}
        \includegraphics[width=\textwidth]{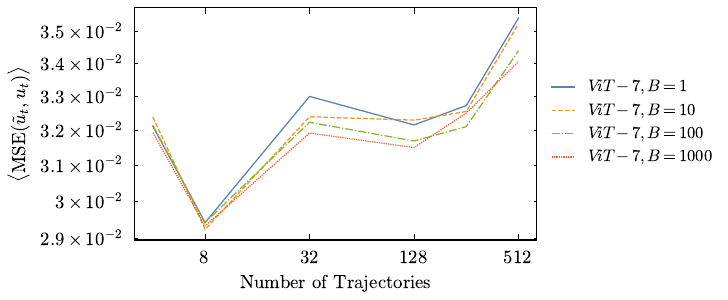}
    \caption{}
        \label{fig:RM_prm}
    \end{subfigure}
    \hfill
    \begin{subfigure}[t]{0.5\textwidth}
        \includegraphics[width=\textwidth]{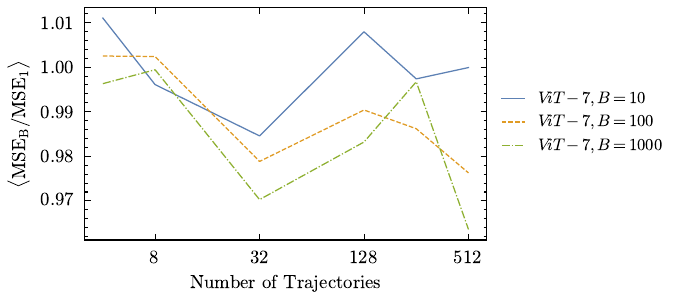}
    \caption{}
        \label{fig:RM_prm_ratio}
    \end{subfigure}
    \hfill
    \begin{subfigure}[t]{0.5\textwidth}
        \includegraphics[width=\textwidth]{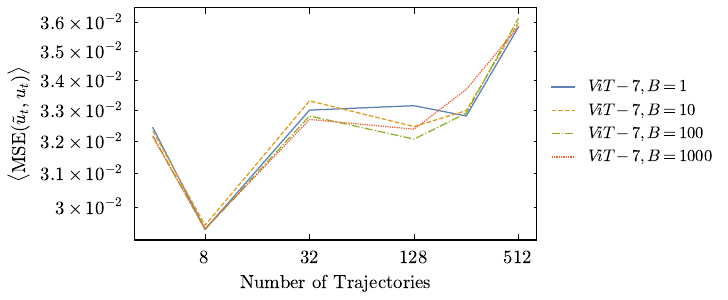}
    \caption{}
        \label{fig:RM_mass}
    \end{subfigure}
    \hfill
    \begin{subfigure}[t]{0.5\textwidth}
        \includegraphics[width=\textwidth]{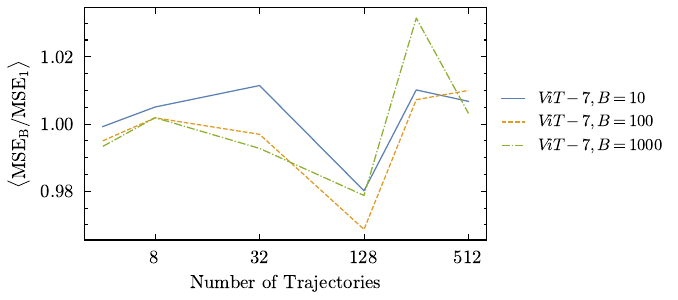}
    \caption{}
        \label{fig:RM_mass_ratio}
    \end{subfigure}
    
    \caption{Performance on RM for PRM (a,b) and Mass Conservation (c,d) reward model}
    \label{fig:RM_composite}
\end{figure}

\section{Finetuning Physical Metrics Measurements}

\begin{figure}[H]
    \begin{subfigure}[t]{0.5\textwidth}
        \includegraphics[width=\textwidth]{boxplot_files/Finetune_Box_prm/RPUI_10_Box_MSE_Patch_3.pdf}
    \caption{}
        \label{fig:RPUI_prm_MSE_Box}
    \end{subfigure}
    \hfill
    \begin{subfigure}[t]{0.5\textwidth}
        \includegraphics[width=\textwidth]{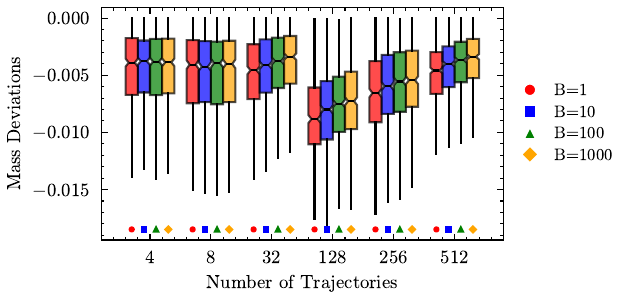}
    \caption{}
        \label{fig:RPUI_prm_Mass_Box}
    \end{subfigure}
    \hfill
    \begin{subfigure}[t]{0.5\textwidth}
        \includegraphics[width=\textwidth]{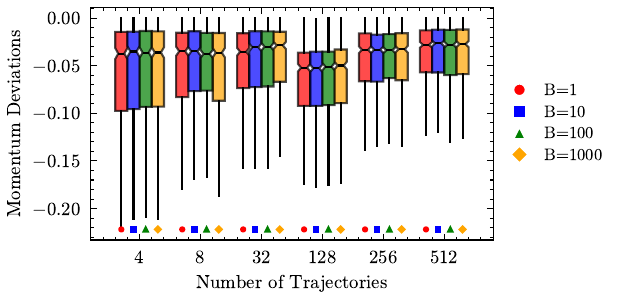}
    \caption{}
        \label{fig:RPUI_prm_Momentum_Box}
    \end{subfigure}
    \hfill
    \begin{subfigure}[t]{0.5\textwidth}
        \includegraphics[width=\textwidth]{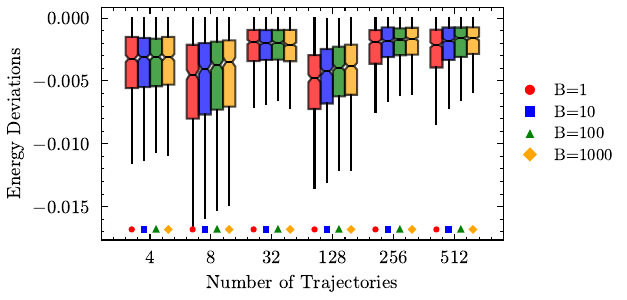}
    \caption{}
        \label{fig:RPUI_prm_Energy_Box}
    \end{subfigure}
    
    \caption{Conservation Metrics on RPUI for PRM Reward model}
    \label{fig:RPUI_Conservation_composite_Box}
\end{figure}

\begin{figure}[H]
    \begin{subfigure}[t]{0.5\textwidth}
        \includegraphics[width=\textwidth]{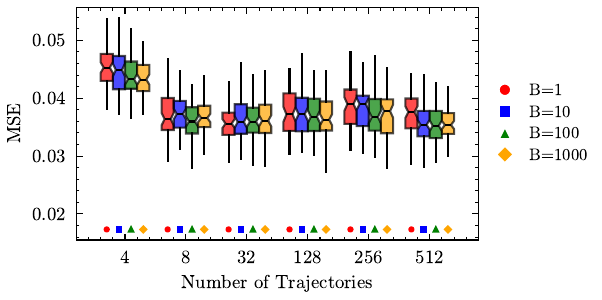}
    \caption{}
        \label{fig:RM_prm_MSE_Box}
    \end{subfigure}
    \hfill
    \begin{subfigure}[t]{0.5\textwidth}
        \includegraphics[width=\textwidth]{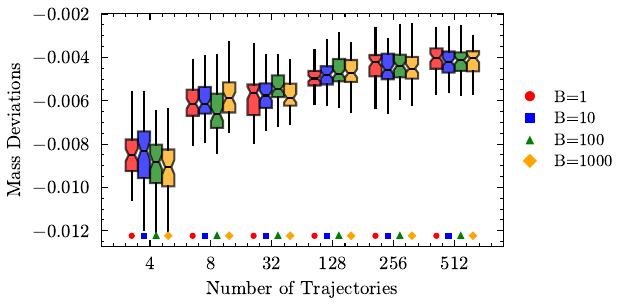}
    \caption{}
        \label{fig:RM_prm_Mass_Box}
    \end{subfigure}
    \hfill
    \begin{subfigure}[t]{0.5\textwidth}
        \includegraphics[width=\textwidth]{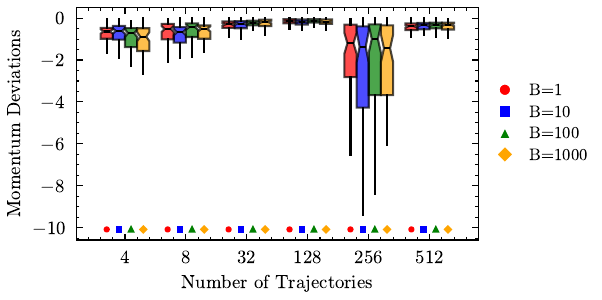}
    \caption{}
        \label{fig:RM_prm_Momentum_Box}
    \end{subfigure}
    \hfill
    \begin{subfigure}[t]{0.5\textwidth}
        \includegraphics[width=\textwidth]{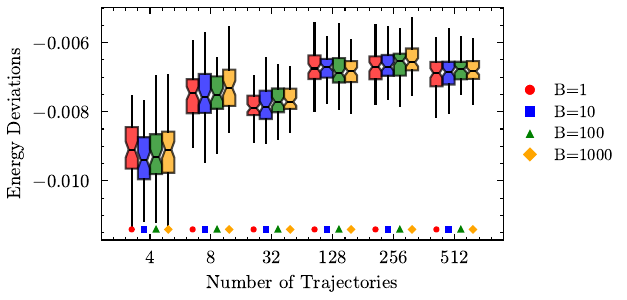}
    \caption{}
        \label{fig:RM_prm_Energy_Box}
    \end{subfigure}
    
    \caption{Conservation Metrics on RM for PRM Reward model}
    \label{fig:RM_Conservation_composite_Box}
\end{figure}
